\documentclass[12pt]{article}
\usepackage{amsmath}
\usepackage{amsthm}
\usepackage{amssymb}
\usepackage{amstext}
\usepackage{algorithm}
\usepackage{algpseudocode}
\usepackage{bm}
\usepackage{booktabs} %toprule bottomrule
\usepackage{color}
\usepackage{enumerate} % must be put before enumitem
\usepackage[shortlabels]{enumitem}
\usepackage{epsfig,epsf,psfrag}
\usepackage{epstopdf}
\usepackage{float}
\usepackage{graphics}
\usepackage{graphicx}
\usepackage[colorlinks,linkcolor=black,citecolor={blue!50!black},urlcolor={blue!50!black}]{hyperref} 
\usepackage{lscape}
\usepackage{mathabx} % for \widecheck
\usepackage{mathtools}
\usepackage{mathrsfs}
\usepackage{multirow}
\usepackage{natbib}
\usepackage{rotate}
\usepackage{subcaption}
\usepackage{url}
\usepackage{xargs}[2008/03/08]
\usepackage{xcolor}
\usepackage{xparse}
\usepackage{xspace}
\usepackage{xr}

\newtheorem{theorem}{Theorem}%[section]

\newtheorem{corollary}[theorem]{Corollary}
{
	\theoremstyle{remark}
	\newtheorem{remark}{Remark}
	
    \newtheorem{assumption}{Assumption}
    
    \newtheorem{case}{Experiment}
}

\newtheorem*{lemma*}{Lemma} % macros for theorem styles

\definecolor{DarkBlue}{rgb}{0,.08,.45}
\definecolor{DarkRed}{rgb}{.7,0,.4}
\definecolor{DarkGreen}{rgb}{0.025,.5,.17}

\newcommand{\bea}{\begin{eqnarray*}}
\newcommand{\eea}{\end{eqnarray*}}
\newcommand{\be}{\begin{eqnarray}}
\newcommand{\ee}{\end{eqnarray}}
\newcommand{\beq}{\begin{equation}}
\newcommand{\eeq}{\end{equation}}

\newcommand{\bgt}{\begin{equation}\begin{gathered}}
\newcommand{\egt}{\end{gathered}\end{equation}}
\newcommand{\bal}{\begin{equation}\aligned}
\newcommand{\eal}{\endaligned\end{equation}}
\newcommand{\ed}{\end{document}}

\newcommand{\btab}{\begin{tabular}}
\newcommand{\etab}{\end{tabular}}

\newcommand{\bi}{\begin{itemize}}
\newcommand{\ei}{\end{itemize}}
\newcommand{\bfi}{\begin{figure}}
\newcommand{\efi}{\end{figure}}
\newcommand{\ben}{\begin{enumerate}}
\newcommand{\een}{\end{enumerate}}
\newcommand{\bay}{\begin{array}}
\newcommand{\eay}{\end{array}}

\newcommand{\nn}{\nonumber}

\newcommand{\bc}{\begin{center}}
\newcommand{\ec}{\end{center}}

\def\Cov{{\rm Cov}}

\def\diag{{\rm diag}}

\def\tps{^{\top}}

\def\expit{\mathrm{expit}}

\def\eps{\varepsilon}

\def\id{{\rm id}}

\def\diffop{\mathrm{d}}

\def\inv{^{-1}}
\def\wh{\widehat}
\def\wt{\widetilde}

\def\ra{\rightarrow}

\def \mbb {\mathbb}
\def \mbf {\mathbf}

\DeclareMathOperator*{\esssup}{ess\,sup}

\DeclareMathOperator*{\argmin}{arg\,min}

\def\unif{\mathrm{Unif}}
\def\bernoulli{\mathrm{Bernoulli}}
\def\normal{\mathcal{N}}
\def\Gammadistn{\mathrm{Gamma}}

\def\E{\mbb{E}} % expectation
\def\prob{\mbb{P}} % probability
\def\real{\mbb{R}}
\def\R{\mathbb{R}}

\def\Op{O_{\prob}}
\def\op{o_{\prob}}
\newcounter{const}
\NewDocumentCommand{\constant}{o}
 {% #1 = number (optional)
  \IfValueTF{#1}%    
  {\ensuremath{C_{#1}}}%
  {\refstepcounter{const}%
  \ensuremath{C_{\theconst}}}%
 }
\newcounter{fun}
\NewDocumentCommand{\function}{oo
}{
  \IfValueTF{#2}%    
  {\ensuremath{f_{\substack{#2,#1}}}}%
  {\refstepcounter{fun}%
  \ensuremath{f_{\substack{\thefun,#1}}}}%
 }
\newcounter{fcl}
\NewDocumentCommand{\fclass}{o}
 {% #1 = (optional) 
  \IfValueTF{#1}%    
  {\ensuremath{\mathcal{F}_{#1}}}%
  {\refstepcounter{fcl}%
  \ensuremath{\mathcal{F}_{\thefcl}}}%
 }
\usepackage[margin=1in,centering]{geometry}
\usepackage{setspace}

\usepackage{sectsty}
\sectionfont{\centering\large}
\subsectionfont{\bf\normalsize}
\subsubsectionfont{\it\normalsize}

\renewenvironment{thebibliography}[1]{
	\begin{oldthebibliography}{#1}
		\setlength{\parskip}{0ex}
		\setlength{\itemsep}{0ex}
	}
	{
	\end{oldthebibliography}
} % styles that possibly need to be eliminated for journal submission

\def\references{\bibliography{merged}}

\graphicspath{{figs/}}

\def\sppl{Supplement\xspace}

\def\actv{\sigma} % activation function
\def\bbN{\mathbb{N}} % set of all nonnegative integers
\def\f{m} % all functions
\def\cfmean{m_\oplus} % conditional \F mean
\def\dEuc{d_{\mathrm{E}}} % Euclidean metric
\def\adfnnbase{aDFNN$_{\mathrm{base}}$\xspace}  % approx loss, small net
\def\adfnn{aDFNN\xspace}                          % approx loss, big net
\def\dfnn{DFNN\xspace}                            % exact loss, big net
\def\dfnnbase{DFNN$_{\mathrm{base}}$\xspace} 

\def\ballatomega{\mathcal{B}^\omega}

\def\dimX{p}
\def\domvX{\mathcal{D}} % domain of random vector \vX
\def\edgewt{E} % for the network data example
\def\F{Fr\'echet\xspace}
\def\funlyr{g} % function mapping input to each layer
\def\funout{\funlyr_{\nlyr+1}} % function mapping input to output prediction
\def\gfrwt{s} % global \F weight function
\def\gfrwthat{\widehat{s}} % estimated global \F weight function
\def\ltwo{\mathcal{L}_2}
\def\linfty{\mathcal{L}_{\infty}}
\def\msp{\Omega} % metric space
\def\mwt{\mathbf{W}} % weight matrices

\def\mask{A}
\def\mA{{\mathbf{A}}}
\def\mD{{\mathbf{D}}}
\def\mE{{\mathbf{E}}}
\def\mG{{\mathbf{G}}}
\def\mH{{\mathbf{H}}}
\def\mI{{\mathbf{I}}}
\def\mL{{\mathbf{L}}}
\def\mP{{\mathbf{P}}}
\def\nlyr{L} % # of layers
\def\nMC{B} % # of MC runs
\def\nepint{N_{\mathrm{initial}}} % epoch_initial
\def\nnode{q}% # of nodes in the network data examples
\def\npatn{N_{\mathrm{patience}}} %\texttt{patience}
\def\ofcfmean{M} % obj. func. for conditional \F mean
\def\ofFNN{M^{\mathrm{NN}}} % obj. func. for the global \F layer
\def\ofFNNhat{\widehat{M}^{\mathrm{NN}}} % empirical obj. func. for the global \F layer
\def\predFNN{m_\oplus^{\mathrm{NN}}} % global \F layer
\def\predFNNhat{\widehat{m}_\oplus^{\mathrm{NN}}} % empirical global \F layer
\def\predY{\wt{Y}} % predicted Y
\def\risk{R}
\def\efrisk{\wh{\risk}}
\def\spredY{\wh{Y}} % sample version of $\predY$
\def\radofcfmean{\rho_{\star}} 
\def\radofpredFNN{\eta_{\star}} 
\def\trwtset{\wtset^*} % true weight set
\def\trwtsethat{\widehat{\wtset}^*} % true weight set
\def\sphere{\mathbb{S}}
\def\predvar{{x}} 
\def\pred{{X}} 
\def\tolr{\delta_{\mathrm{tolerance}}} %min_delta
\def\uppwtop{W^\star}

\def\uppshift{B^\star}
\def\uppwtfrob{W^{0}}
\def\va{{\bm{a}}}
\def\vb{{\bm{b}}}
\def\ve{{\bm{e}}}
\providecommand\vx{{\bm{\predvar}}} %vector x
\def\vq{{\bm{q}}}
\def\vv{{\bm{v}}}
\def\vshift{{\bm{b}}} 
\def\vu{{\bm{u}}}
\def\vz{{\bm{z}}}
\def\vX{{\bm{\pred}}} %random vector X
\def\veps{\eps'}
\def\wtset{{\bm{\theta}}} % \mwt and \vshift
\def\wtdom{\Theta} % domain of $\wtset$
\def\wlyr{p} % width of $l$-th layer: $\wlyr_l$
\def\wout{\wlyr_{\nlyr+1}} % width of output layer

\def\flyrMean{\bm{\mu}_{\wtset_\nlyr}} %mean of the final layer
\def\flyrCov{\mathbf{\Sigma}_{\wtset_\nlyr}} %covariance of the final layer
\def\flyrMeanhat{\widehat{\bm{\mu}}_{\wtset_\nlyr}} %mean of the final layer
\def\flyrCovhat{\widehat{\mathbf{\Sigma}}_{\wtset_\nlyr}} %covariance of the final layer
\def\nlyrapprox{\nlyr_{\delta}} % number of hidden layers $\nlyr$ related to approximation level $\delta$
\def\wtsetapprox{\wtset_{\nlyrapprox,\delta}} % $\wtset$ related to approximation level $\delta$
\def\sblevsymb{\mathcal{W}} %symbol used for the Sobolev space
\def\sblevorder{r}% order of Sobolev space 
\def\sblevordofcfm{\sblevorder_{\ofcfmean}}% order of Sobolev space including the objective function of conditional \F mean
\def\sblevorderdiv{\nu}% breakdown of order of Sobolev space
\def\sblev{\sblevsymb^{\sblevorder,\infty}(\domvX)} %sobolev space
\def\sbradius{K} %radius of sobolev space
\def\sbradofcfm{{\sbradius_{\ofcfmean}}} %radius of sobolov ball including the objective function of conditional \F mean
\def\sblevcmpkt#1#2{\sblevsymb^{#2,\infty}(\domvX, #1)} % compact sobolev space
\def\oprt{\mathrm{op}} % operator norm
\def\frob{\mathrm{F}}  % frobenius norm
\renewcommand{\tilde}{\widetilde}
\renewcommand{\hat}{\widehat} % macros for notations specific to this paper

\usepackage{authblk}

\title{DFNN: A Deep Fr\'echet Neural Network Framework for Learning Metric-Space-Valued Responses\thanks{This research was supported in part by NSF grants DMS-2311035 (YC) and DMS-2311034 (PD).}}
\author[1]{Kyum Kim}
\author[1]{Yaqing Chen\footnote{Corresponding authors.}}
\newcommand\CoAuthorMark{\footnotemark[\arabic{footnote}]}
\author[2]{Paromita Dubey\protect\CoAuthorMark}
\affil[1]{Rutgers University}
\affil[2]{University of Southern California}
\date{}

\begin{document}

\maketitle

\begin{abstract}
Regression with non-Euclidean responses---e.g., probability distributions, networks, symmetric positive-definite matrices, and compositions---has become increasingly important in modern applications. In this paper, we propose deep Fr\'echet neural networks (DFNNs), an end-to-end deep learning framework for predicting non-Euclidean responses---which are considered as random objects in a metric space---from Euclidean predictors. Our method utilizes the representation-learning power of deep neural networks (DNNs) to the task of approximating conditional Fr\'echet means of the response given the predictors, the metric-space analogue of conditional expectations, by minimizing a Fr\'echet risk. The framework is highly flexible, accommodating diverse metrics
and high-dimensional predictors. We establish a universal approximation theorem for DFNNs, advancing the state-of-the-art of neural network approximation theory to general metric-space-valued responses, without making model assumptions or relying on local smoothing. We further establish rigorous generalization guarantees for DFNNs and derive corresponding risk bounds, providing, to the best of our knowledge, the first such theoretical results for deep learning regression with metric-space-valued responses. Empirical studies on synthetic distributional and network-valued responses, as well as real-world applications to predicting compositional responses in an Aitchison simplex and spherical responses, demonstrate that DFNNs consistently outperform all existing methods.
\end{abstract}


\section{Introduction}\label{sec:intro}

Regression and supervised learning are central topics in statistics and machine learning. Modern applications increasingly involve non-Euclidean responses---e.g., probability distributions, networks, symmetric positive-definite (SPD) matrices, and compositional data---that reside in nonlinear metric spaces rather than unconstrained Euclidean spaces, paired with Euclidean covariates/features. These settings arise across a wide range of applications, including brain connectivity networks and diffusion tensors in neuroscience \citep{fair:08,benn:10}, microbial compositions in microbiome studies \citep{turn:09}, income distributions in economics \citep{jant:10}, and single-cell gene expression distributions in drug studies \citep{aiss:21}. Such data, often considered as random objects on metric spaces and also referred to as object data and object-oriented data \citep{marr:14,mull:16:9,marr:21,dube:24,song:26}, lack the vector-space structure underlying classical methods, posing fundamental challenges to learning metric-space-valued responses---in particular, even the notion of means is no longer well-defined. Consequently, conventional regression methods cannot be applied directly, motivating the development of new methods for predicting metric-space-valued responses from Euclidean covariates \citep{mull:16:9}.

In statistics, several approaches have been developed for regression with non-Euclidean responses. Early extrinsic methods embed non-Euclidean objects into Euclidean spaces, often through pairwise distances, and then apply conventional regression techniques \citep{fara:14}. Intrinsic approaches instead operate directly on the geometry of the response space, including Nadaraya--Watson-type kernel smoothing \citep{hein:09} and \F regression \citep{pete:19,chen:22,scho:22}. Among the latter, global \F regression extends multiple linear regression to metric-space-valued responses without smoothing or tuning parameters, but relies on assumptions analogous to classical multiple linear models. Local \F regression, a generalization of local linear smoothing to metric-space-valued responses,  is more flexible but suffers from the curse of dimensionality. Overall, extrinsic methods, while simple, may distort the underlying geometry of the metric space, while intrinsic methods can be limited by restrictive model assumptions or poor scalability to moderately high-dimensional predictors.

To accommodate high-dimensional predictors, recent work has adapted classical dimension reduction methods to \F regression, including principal component regression \citep{song:23:2}, single-index models \citep{ghos:23:2,bhat:23:3}, and sufficient dimension reduction \citep{dong:22,virt:22,ying:22,zhan:24:2,weng:25,ying:25}. These approaches, however, often rely on restrictive linearity or distributional assumptions, require tuning, or require careful construction of kernels. More recently, \cite{iao:25} proposed a deep learning approach that learns a low-dimensional manifold representation of the regression function and applies local \F regression on this representation. This method assumes that the regression function lies on a manifold of intrinsic dimension $r$ and remains subject to the curse of dimensionality of local \F regression, becoming impractical as $r$ grows beyond $3$.

To overcome these limitations, we propose a deep \F neural network (deep FNN or DFNN) framework for directly predicting metric-space-valued responses from potentially high-dimensional Euclidean predictors. DFNN targets the conditional \F mean \citep{frec:48}, the metric-space analogue of the conditional mean which is a natural regression target in this setting. Its core innovation lies in the output layer, while the input and hidden layers retain the standard classical DNN architecture. Specifically, the output layer consists of a single metric-space-valued node defined, as per Eq.~\eqref{eq:dfnn_output_layer}, as a weighted \F mean of the responses, with weights determined by the final hidden-layer representation; consequently, DFNN predictions remain in the underlying response space. In the Euclidean case, this output layer reduces to the least-squares predictor obtained by regressing the response on the final hidden-layer features. 

DFNN is trained by minimizing the \emph{\F risk}, defined as the mean squared distance between the responses and their predictions. For Euclidean responses, we show that DFNN training is equivalent to training a standard DNN with the same hidden-layer architecture (Remark~\ref{rem: dfnn_to_dnn}). We establish a universal approximation theorem for the proposed DFNN framework (Theorem~\ref{thm:approximation_theorem}), showing that DFNNs can approximate the conditional \F mean arbitrarily well and quantify how this approximation error translates into excess \F risk and prediction error (Corollary~\ref{cor: pred_approx}), thereby providing the theoretical foundation for learning metric-space-valued responses from Euclidean predictors.

Beyond universal approximation, we develop a statistical theory for DFNNs trained by empirical \F risk minimization. Specifically, we establish uniform consistency and convergence rates for the empirical DFNN output layer over the predictor domain and neural network parameter space (Theorem~\ref{thm: output_layer_convergence}). Building on these results, we derive generalization guarantees for the empirical \F risk minimizer, with explicit bounds on its excess \F risk and prediction error that separate estimation, neural network complexity, and approximation errors (Theorem~\ref{thm: risk_and_pred_error_bound}). These bounds yield concrete rates under both norm-based and parameter-count-based complexity controls of the neural network architecture (Remarks~\ref{rem: entropy_rate} and \ref{rem: entropy_rate_fast}). To the best of our knowledge, these provide the first generalization guarantees for deep learning with responses in general metric spaces. 

Recently, \cite{zhou:26:2} independently proposed a deep neural network framework for regression with metric-space-valued responses and Euclidean predictors. 
However, their framework is different from ours, in particular in the way how predictions of responses are structured in the output layer. 
Moreover, the assumptions underlying their framework are also distinct from ours---e.g., they assume the conditional \F mean takes the form of a weighted \F mean with the weights given by the last hidden layer of the network, which are nonnegative and summed up to $1$, and that the metric space is compact. 
Neither of these assumptions is required in the DFNN framework we propose. 
Overall, both the methodology and the theory developed in this paper are fundamentally different from those in \cite{zhou:26:2}. 

To validate the effectiveness of DFNNs, we conducted extensive experiments on synthetic data simulated in three different settings, 
including distributional responses, 
and network-valued responses, 
as well as real-world applications of predicting compositional responses in an Aitchison simplex and spherical responses from U.S.\ states' employment occupational data and electricity generation data using economic, geographic, and demographic features, covering non-Euclidean responses in diverse metric spaces. 
Results show that DFNN constantly and significantly outperforms the existing methods across responses in different metric spaces. 

The remainder of the paper is organized as follows. In the rest of Section~\ref{sec:intro}, we highlight the main contributions of this work. More detailed review of related work is outlined in Section~\ref{sec:related} of the \sppl. Section~\ref{sec:method} introduces the proposed DFNN framework for regression with metric-space-valued responses and Euclidean predictors. Section~\ref{sec:theory} develops the theoretical properties of DFNNs, including universal approximation, uniform consistency and convergence rates, and generalization guarantees with corresponding risk and prediction-error bounds. 
Section~\ref{sec:empircal} presents implementation details and empirical studies demonstrating the prediction performance of DFNNs across different types of non-Euclidean responses.

\subsection*{Main Contributions}
This paper develops a deep learning framework and corresponding theoretical foundations for regression with metric-space-valued responses and Euclidean predictors. The main contributions are fourfold.
\begin{itemize}
\item We propose an end-to-end deep \F neural network (DFNN) framework for regression with metric-space-valued responses and Euclidean predictors, which extends classical DNNs to the scope of predicting non-Euclidean responses.  The DFNN framework is highly flexible, capable of accommodating diverse metric spaces and high-dimensional predictors without restricting to shallow models or parametric/semiparametric models or relying on any low-dimensional (manifold) embeddings of the regression target. 
The prediction of non-Euclidean responses obtained by training a DFNN respects the geometry of the response space. 
\item We establish a universal approximation theorem showing that DFNNs can approximate the conditional \F mean of the response given the predictors---the natural regression target for metric-space-valued responses---arbitrarily well. This extends neural network approximation theory to responses taking values in general metric spaces.
\item We develop statistical theory for DFNNs trained by empirical \F risk minimization. We establish uniform consistency and convergence rates for the empirical DFNN output layer over the predictor domain and neural network parameter space, and derive generalization guarantees for the empirical \F risk minimizer through bounds on its excess \F risk and prediction error. We further provide explicit rates under both norm-based and parameter-count-based complexity controls. To the best of our knowledge, these are the first such generalization results for deep learning with responses in general metric spaces.
\item We conduct extensive experiments on synthetic and real data with distributional, network-valued, and compositional responses. Across the settings considered, DFNN consistently achieves superior prediction performance compared with existing competing methods for regression with non-Euclidean responses.
\end{itemize}

\section{DFNN: Deep \F Neural Network}\label{sec:method}

Let $(\msp,d)$ be a separable metric space. 
Consider a random pair $(\vX,Y)$ with a joint distribution on the product space $\domvX\times\msp$, where $\domvX$ is a compact subset of $\R^{\dimX}$, without loss of generality assumed to be $[0,1]^{\dimX}$ in this paper. 
In regression tasks, the goal is to estimate the conditional \F mean of $Y$ given $\vX = \vx$, denoted by $\cfmean(\vx)$ for any $\vx\in\domvX$; specifically, $\cfmean(\vx)$ is defined as 
\bal\label{eq:cfmean}
\cfmean(\vx) \in \argmin_{\omega\in\msp} \ofcfmean(\omega,\vx)
\quad \text{with}\ 
\ofcfmean(\omega,\vx) = \E\{d^2(Y,\omega) \mid \vX=\vx\},
\eal
where we assume that for each $\vx \in \domvX$, there exists $\omega_{\vx} \in \Omega$ such that $\ofcfmean(\omega_{\vx},\vx) < \infty$. To this end, we propose a deep \F neural network (DFNN) framework which integrates global \F regression \citep{pete:19} with deep neural network structures and provides an end-to-end regression method for metric-space-valued responses paired with Euclidean predictors. Specifically, we define a DFNN with $\nlyr$ hidden layers as follows.

Let $\wlyr_0 =\dimX$. 
For each $l=1,\dots,\nlyr$, 
suppose the $l$-th hidden layer consists of $\wlyr_{l}$ nodes.  
Let $\mwt_{l}$ be a weight matrix in $\R^{\wlyr_{l}\times\wlyr_{l-1}}$, $\vshift_{l}\in\R^{\wlyr_{l}}$ be a shift vector, and
$\wtset_{l} = (\mwt_{1},\vshift_{1},\mwt_{2},\vshift_{2},\dots,\newline\mwt_{l},\vshift_{l})$ be the weight matrices $\{\mwt_{k}\}_{k=1}^{l}$ and shift vectors $\{\vshift_{k}\}_{k=1}^{l}$ of the first $l$ hidden layers, 
which takes values in $\wtdom_{l} = \R^{\wlyr_{1}\times\wlyr_{0}}\times \R^{\wlyr_{1}}\times \R^{\wlyr_{2}\times\wlyr_{1}}\times \R^{\wlyr_{2}}\times\dots \times \R^{\wlyr_{l}\times\wlyr_{l-1}}\times \R^{\wlyr_{l}}$. 
Let $\funlyr_{l}(\,\cdot\,;\wtset_{l}) \colon \R^{\dimX}\ra\R^{\wlyr_{l}}$ denote the map that maps the input layer of $\dimX$ predictors to the $l$-th hidden layer of $\wlyr_{l}$ nodes. 
For any given $\{\mwt_{l},\vshift_{l}\}_{l=1}^{\nlyr}$, the maps $\{\funlyr_{l}(\,\cdot\,;\wtset_{l})\}_{l=1}^{\nlyr}$ can be defined in a recursive way. 
As an initialization, define $\wtset_{0} = \emptyset$ and 
$\funlyr_0(\,\cdot\,;\wtset_{0})=\id(\cdot)$ being the identity function, i.e., $\funlyr_0(\vx;\wtset_{0}) = \vx$ for all $\vx\in\R^{\dimX}$. 
Then for each $l=1,\dots,\nlyr$, 
the map $\funlyr_{l}(\,\cdot\,;\wtset_{l})$ is defined as 
\bal\nn
\funlyr_{l}(\vx;\wtset_{l}) = \actv\{\mwt_{l}\funlyr_{l-1}(\vx;\wtset_{l-1}) + \vshift_{l}\} 
\quad\text{for } \vx\in\R^{\dimX},
\eal
where $\actv(\cdot)$ denotes the component-wise rectifier linear unit (ReLU) activation function, i.e., $\actv(\va) = (\max\{0,a_{1}\},\dots,\max\{0,a_{q}\})\tps$ for $\va\in\R^{q}$ for any integer $q\ge 1$. 
The output layer of a DFNN is defined as a layer of a single node, denoted by $\predFNN(\vx;\wtset_{\nlyr})$, which serves as the prediction of the response: 
\bal\label{eq:dfnn_output_layer}
\predFNN(\vx;\wtset_{\nlyr})
&\in \argmin_{\omega\in\msp} \ofFNN(\omega,\vx;\wtset_{\nlyr}) 
\\ \text{with } 
\ofFNN(\omega,\vx;\wtset_{\nlyr})
&=\E\left[\gfrwt\{\funlyr_{\nlyr}(\vx;\wtset_{\nlyr}),\funlyr_{\nlyr}(\vX;\wtset_{\nlyr})\} d^2(Y,\omega) \right].
\eal
Here, for any $\vx,\vx'\in\domvX$, 
\bal\label{eq:dfnn_output_weight}
\gfrwt\{\funlyr_{\nlyr}(\vx;\wtset_{\nlyr}),\funlyr_{\nlyr}(\vx';\wtset_{\nlyr})\} = 1 + (\funlyr_{\nlyr}(\vx;\wtset_{\nlyr}) - \flyrMean )\tps \flyrCov\inv(\funlyr_{\nlyr}(\vx';\wtset_{\nlyr}) - \flyrMean ),
\eal
with $\flyrMean = \E\{\funlyr_{\nlyr}(\vX;\wtset_{\nlyr})\}$ and $\flyrCov = \Cov\{\funlyr_{\nlyr}(\vX;\wtset_{\nlyr})\}$, 
corresponds to the global \F regression weights \citep{pete:19} applied to the final hidden-layer representations $\funlyr_{\nlyr}(\vx;\wtset_{\nlyr})$, 
and $\ofFNN(\omega,\vx;\wtset_{\nlyr})$ is the global \F regression objective function. At the population level, a DFNN is trained by minimizing the \F risk of $\predFNN(\cdot;\wtset_{\nlyr})$. 
Specifically, for any predictor $\f\colon\R^\dimX \to \Omega$, its \F risk is defined by 
\bal \label{eq:risk}
\risk(\f) = \E_{Y,\vX}\{d^2[Y,\f(\vX)]\}, 
\eal
which measures the mean squared distance between the response $Y$ and the prediction $\f(\vX)$. 
Hence, the \F risk of $\predFNN(\cdot;\wtset_{\nlyr})$ is given by 
\bal\label{eq:dfnn_risk_pop}
\risk\{\predFNN(\cdot;\wtset_{\nlyr})\} 
= \E\{d^2[Y,\predFNN(\vX;\wtset_{\nlyr})]\}.
\eal
Let $\trwtset_{\nlyr}
\in \argmin_{\wtset_{\nlyr}\in\wtdom_{\nlyr}} \risk\{\predFNN(\cdot;\wtset_{\nlyr})\}$ be a (population) \emph{\F risk minimizer}, which is assumed to exist in the parameter space $\wtdom_{\nlyr}$.    
A trained DFNN predicts the response $Y$ given $\vX=\vx$ by 
\bal\label{eq:dfnn_prediction_pop}
\predY(\vx) 
= \predFNN(\vx;\trwtset_{\nlyr}) 
\quad\text{for any}\ \vx\in\R^{\dimX}.
\eal

\begin{figure}[!hbt]
\centering
\includegraphics[width=0.90\linewidth]{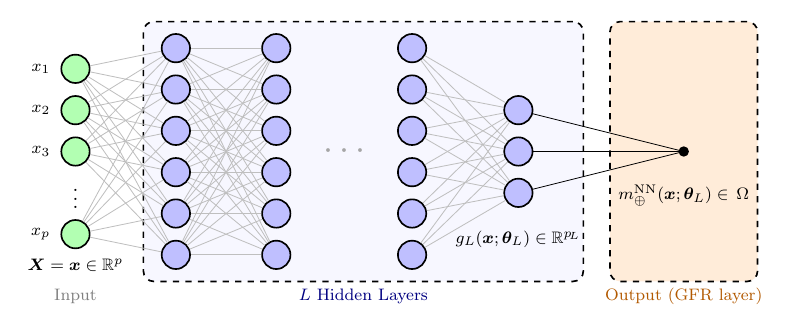}
\vspace{-.5em}
\caption{Architecture of a deep Fr\'echet neural network (DFNN). The input $\vX=\vx\in\R^{\dimX}$ is passed through $\nlyr$ hidden layers to produce an embedding $\funlyr_{\nlyr}(\vx;\wtset_{\nlyr})\in\R^{\wlyr_{\nlyr}}$, which is then fed into the global Fr\'echet regression (GFR) layer to produce a prediction $\predFNN(\vx;\wtset_{\nlyr})\in\msp$.}
\label{fig:dfnn_arch}
\end{figure}

\begin{remark}
\label{rem: dfnn_to_dnn}
For $\msp=\R$ with the Euclidean metric $d=\dEuc$, minimizing the \F risk in Eq.~(\ref{eq:dfnn_risk_pop}) for the DFNN output $\predFNN(\vx;\wtset_{\nlyr})$ is equivalent to minimizing the mean-squared error of a standard DNN with $\nlyr$ hidden layers and a linear activation in the output layer. 
Consider a DNN with $\nlyr$ hidden layers, a single output node ($\wout=1$), and parameters $\wtset_{\nlyr+1}=(\mwt_{1},\vshift_{1},\dots,\mwt_{\nlyr+1},\vshift_{\nlyr+1})$. In a standard DNN, with $\mwt_{\nlyr+1}\in\R^{1\times \wlyr_{\nlyr}}$ and $\vshift_{\nlyr+1}\in\R$, the output node is given by 
\bal\nn
\funout(\vx;\wtset_{\nlyr+1}) = \mwt_{\nlyr+1}\funlyr_{\nlyr}(\vx;\wtset_{\nlyr}) + \vshift_{\nlyr+1}
\quad\text{for } \vx\in\R^{\dimX}.
\eal
At the population level, training a DNN by minimizing the mean squared error over $\wtset_{\nlyr+1}$ can be expressed as 
\bal\label{eq:dnn_model}
&\min_{\wtset_{\nlyr+1}} \E \{Y-\funout(\vX;\wtset_{\nlyr+1})\}^2
\\&= \min_{\wtset_{\nlyr+1}} \E\{Y-\mwt_{\nlyr+1}\funlyr_{\nlyr}(\vx;\wtset_{\nlyr}) - \vshift_{\nlyr+1}\}^2
\\&=\min_{\wtset_{\nlyr}} \min_{\mwt_{\nlyr+1},\vshift_{\nlyr+1}} 
\E\left[\dEuc^2\{Y, \mwt_{\nlyr+1}\funlyr_{\nlyr}(\vx;\wtset_{\nlyr}) + \vshift_{\nlyr+1}\} \right]
\\&= \min_{\wtset_{\nlyr}} \E\left(\dEuc^2\left\{Y, \argmin_{\omega\in\msp} \E\left[\gfrwt\{\funlyr_{\nlyr}(\vx;\wtset_{\nlyr}),\funlyr_{\nlyr}(\vX;\wtset_{\nlyr})\} \dEuc^2(Y,\omega) \right]\right\}\right) 
\\&= \min_{\wtset_{\nlyr}} \E\left(\dEuc^2\left\{Y, \predFNN(\vX;\wtset_{\nlyr})\right\}\right),
\eal
where the third equality in Eq.~\eqref{eq:dnn_model} is implied by the global \F regression formulation in Section~2.2 of \cite{pete:19}, with $\gfrwt\{\funlyr_{\nlyr}(\vx;\wtset_{\nlyr}),\funlyr_{\nlyr}(\vX;\wtset_{\nlyr})\}$ given in Eq.~(\ref{eq:dfnn_output_weight}). Thus, in the Euclidean case, DFNN training by minimizing \F risk coincides with standard DNN training by minimizing mean squared error.

\end{remark}

Remark~\ref{rem: dfnn_to_dnn} shows that for scalar responses with the Euclidean metric---i.e., $\msp=\mathbb{R}$ and $d=\dEuc$---the DFNN predictor coincides with that of a standard DNN of the same architecture with a linear activation for the output layer  trained by minimizing mean squared error. Thus, the DFNN framework directly generalizes standard DNNs from scalar responses to metric-space-valued responses. 

In practice, consider a random sample of $n$ i.i.d. pairs $(\vX_1, Y_1), \dots, (\vX_n,Y_n)$ from the joint distribution of $(\vX, Y)$ on the $\domvX\times\msp$, which are independent of $(\vX, Y)$. 
We define the \emph{empirical \F risk} as
\bal\label{eq:dfnn_risk_emp}
\efrisk\{\predFNNhat(\cdot;\wtset_{\nlyr})\} 
=
\frac{1}{n} \sum_{i=1}^n d^2\{Y_i,\predFNNhat(\vX_i;\wtset_{\nlyr})\}
\eal
where for any $\vx \in \R^{\dimX}$, the empirical estimator of the DFNN output layer, $\predFNN(\vx;\wtset_{\nlyr})$, is given by:
\bal\label{eq:dfnn_output_layer_emp}
\predFNNhat(\vx;\wtset_{\nlyr})
&\in\argmin_{\omega\in\msp} \ofFNNhat(\omega,\vx;\wtset_{\nlyr}) 
\\ \text{with } 
\ofFNNhat(\omega,\vx;\wtset_{\nlyr})
&=\frac{1}{n} \sum_{i=1}^n {\gfrwthat}\{\funlyr_{\nlyr}(\vx;\wtset_{\nlyr}),\funlyr_{\nlyr}(\vX_i;\wtset_{\nlyr})\} d^2(Y_i,\omega),
\eal
where for any $\vx,\vx'\in\domvX$, the estimated weights are  
\bal\label{eq:dfnn_output_weight_emp}
\gfrwthat\{\funlyr_{\nlyr}(\vx;\wtset_{\nlyr}),\funlyr_{\nlyr}(\vx';\wtset_{\nlyr})\} 
= 1 + (\funlyr_{\nlyr}(\vx;\wtset_{\nlyr}) - \flyrMeanhat )\tps \flyrCovhat\inv(\funlyr_{\nlyr}(\vx';\wtset_{\nlyr}) - \flyrMeanhat ),
\eal
with $\flyrMeanhat = \frac{1}{n} \sum_{i=1}^n\funlyr_{\nlyr}(\vX_i;\wtset_{\nlyr})$ and $\flyrCovhat = \frac{1}{n} \sum_{i=1}^n\{\funlyr_{\nlyr}(\vX_i;\wtset_{\nlyr})-\flyrMeanhat\}\{\funlyr_{\nlyr}(\vX_i;\wtset_{\nlyr})-\flyrMeanhat\}^\top$. 
Finally, the trained DFNN gives the empirical predictor $\spredY(\cdot)$ given by
\bal\label{eq:dfnn_prediction_emp}
\spredY(\vx) 
= \predFNNhat(\vx;\trwtsethat_{\nlyr}) 
\quad\text{for any}\ \vx\in\R^{\dimX},
\eal
where 
\bal\nn
\trwtsethat_{\nlyr} \in \argmin_{\wtset_{\nlyr}\in\wtdom_{\nlyr}} \efrisk\{\predFNNhat(\cdot;\wtset_{\nlyr})\}
\eal
is an \emph{empirical \F risk minimizer} over the parameter space $\wtdom_{\nlyr}$, which is assumed to exist, and the empirical \F risk $\efrisk$ is defined in Eq.~\eqref{eq:dfnn_risk_emp}. 

\begin{algorithm}[!hbt]
\caption{DFNN Training Procedure}
\label{alg:train_dfnn}
\begin{algorithmic}
\State \textbf{Input:} Training samples $\{(\vX_i, Y_i)\}_{i=1}^n$.
\State \textbf{Output:} Trained neural network parameters $\trwtsethat_{\nlyr}$.
\State \textbf{1:} Initialize network parameters $\wtset_{\nlyr}$.
\State \textbf{2:} Obtain $\trwtsethat_{\nlyr}$ using the stochastic gradient descent (SGD) optimizer to minimize the \emph{empirical \F risk} $\efrisk\{\predFNNhat(\cdot;\wtset_{\nlyr})\}$ defined in Eq.~\eqref{eq:dfnn_risk_emp}.
\end{algorithmic}
\end{algorithm}

In summary, the DFNN framework provides a flexible, model-agnostic way to learn conditional \F means in metric spaces by minimizing the empirical \F risk in Eq.~\eqref{eq:dfnn_risk_emp}. 
While the preceding construction  defines the neural network architecture along with its population and empirical objectives, a central theoretical question remains: 
\emph{How well can DFNN approximate the true conditional \F mean $\cfmean(\cdot)$?} 
Addressing this requires formally analyzing the approximation power of DFNNs for metric space-valued responses. 

In the next section, we establish a universal approximation theorem that guarantees this capability under suitable regularity conditions. Furthermore, we develop the statistical theory of DFNNs, establishing uniform consistency and rates of convergence of the DFNN output layers over the neural network parameter space. 
Finally, we analyze the statistical performance of the empirical DFNN predictor in Eq.~\eqref{eq:dfnn_prediction_emp} obtained by empirical \F risk minimization, providing generalization guarantees through asymptotic bounds in probability on its excess \F risk and prediction error.

\section{Theoretical Results}\label{sec:theory}

With the DFNN architecture and its population and empirical training objectives defined in Section~\ref{sec:method}, we next analyze the theoretical expressive power of this framework. In particular, we show that DFNNs can approximate the conditional \F mean $\cfmean(\cdot)$ defined in Eq.~\eqref{eq:cfmean} arbitrarily well under mild regularity conditions. To specify the smoothness properties of $\ofcfmean(\omega,\vx)$ defined in Eq.~\eqref{eq:cfmean}, we consider the Sobolev spaces $\sblev$ with $\sblevorder = 1,2, \dots$ where $\sblev$ is defined as functions on $\domvX$ lying in $\linfty(\domvX)$ with $\linfty(\domvX)=\{f:\domvX \rightarrow \R\mid \|f\|_\infty < \infty\}$ and $\|f\|_\infty = \esssup_{\vx \in \domvX} |f(\vx)|$, along with their weak derivatives up to order $\sblevorder$. The Sobolev norm of a function $f \in \sblev$ is defined as
\bal\nn
    \| f \|_{\sblev} = \max_{\bm \sblevorderdiv\in\bbN^{\dimX},\, |\bm \sblevorderdiv| \leq \sblevorder} \esssup_{\vx \in \domvX} |D^{\bm \sblevorderdiv} f(\vx)|,
\eal
where $\bbN= \{0,1,2,\dots\}$, $|\bm \sblevorderdiv| =  \sblevorderdiv_1 + \dots + \sblevorderdiv_\dimX$ for $\bm \sblevorderdiv = ( \sblevorderdiv_1, \dots, \sblevorderdiv_{\dimX})\tps \in \bbN^\dimX$, and $D^{\bm \sblevorderdiv} f(\vx) = \frac{\partial^{|\bm \sblevorderdiv|}f(\vx)}{\partial\predvar_1^{\sblevorderdiv_1} \cdots \partial\predvar_{\dimX}^{\sblevorderdiv_{\dimX}}}$. 
The space $\sblev$ can be equivalently described as consisting of functions in $C^{\sblevorder-1}(\domvX)$ such that all their derivatives of order $(\sblevorder-1)$ are Lipschitz continuous. 
For any $\sbradius>0$, let $\sblevcmpkt{\sbradius}{\sblevorder}$ denote the ball centering at $0$ of radius $\sbradius$ in $\sblev$, i.e., 
\bal\label{eq:sblevcmpkt}
\sblevcmpkt{\sbradius}{\sblevorder} = \{ f \in \sblev : \| f \|_{\sblev} \leq \sbradius \}.\eal To establish the universal approximation capability of DFNNs with respect to the conditional \F mean $\cfmean(\cdot)$, we require the following assumptions.

\begin{assumption}
\label{ass:existence}
For any $\vx \in \R^{\dimX}$, the conditional \F mean $\cfmean(\vx)$ exists. Moreover, for every $\vx \in \R^{\dimX}$ and $\wtset_{\nlyr} \in \wtdom_{\nlyr}$, the minimizers $\predFNN(\vx;\wtset_{\nlyr})$ and $\predFNNhat(\vx;\wtset_{\nlyr})$ defined in Eqs.~\eqref{eq:dfnn_output_layer} and \eqref{eq:dfnn_output_layer_emp} exist. 
\end{assumption}

\begin{assumption}
\label{ass:positive_definite_flyr_covariance}
The matrix $\flyrCov$ defined right after Eq.~\eqref{eq:dfnn_output_weight} is positive definite for any $\wtset_{\nlyr}\in\wtdom_{\nlyr}$. 
\end{assumption}

\begin{assumption}
\label{ass:sblev_obj}
There exists an integer $\sblevordofcfm \geq 1$ and a constant $\sbradofcfm > 0$ such that $\ofcfmean(\omega, \cdot) \in \sblevcmpkt{\sbradofcfm}{\sblevordofcfm}$ for all $\omega\in\msp$.
\end{assumption}

\begin{assumption}
\label{ass:local_separation}
There exists $\radofcfmean>0$ such that for any $\rho\in(0,\radofcfmean]$, 
\bal\nn
   \Delta(\rho)=\inf_{\vx \in \domvX}\inf_{d\{\omega,\cfmean(\vx)\}\geq \rho} \{ \ofcfmean(\omega,\vx) - \ofcfmean(\cfmean(\vx),\vx)\} > 0.
\eal 
\end{assumption}

\begin{assumption}
\label{ass:cov_regularity}
There exists a constant $\lambda_{\star}>0$ such that for any $\nlyr\ge 1$ and $\wtset_{\nlyr}\in\wtdom_{\nlyr}$, $\lambda_{\mathrm{min}}(\flyrCov) \geq \lambda_{\star}$, where $\lambda_{\mathrm{min}}(\flyrCov)$ denotes the smallest eigenvalue of the matrix $\flyrCov$.
\end{assumption}

\begin{assumption}
\label{ass:weight_regularity}
There exist constants $\uppwtop\ge 1$ and $ \uppshift > 0$ 
such that $\wtset_{\nlyr}$ satisfies $\prod_{j=l}^{\nlyr} \|\mwt_{j}\|_{\oprt} \leq \uppwtop$ for any $l=1,\dots,\nlyr$ 
and $\max_{1\le l\le\nlyr}\|\vshift_{l}\|\le \uppshift$ for any integer $\nlyr\ge 1$.
\end{assumption}

Assumption~\ref{ass:existence} guarantees that the DFNN output, a global \F regression evaluated at the final hidden layer representation, is well defined for every predictor input and parameter choice. 
Assumption~\ref{ass:positive_definite_flyr_covariance} guarantees that the final layer covariance $\flyrCov$ is nondegenerate, so $\flyrCov\inv$ exists and the weight
$\gfrwt\{\funlyr_{\nlyr}(\vx;\wtset_{\nlyr}),\funlyr_{\nlyr}(\vX;\wtset_{\nlyr})\}$ in Eq.~\eqref{eq:dfnn_output_weight} is well defined for each $\vx\in\R^{\dimX}$ almost surely.
Assumption~\ref{ass:sblev_obj} requiring $\ofcfmean(\omega,\cdot)\in \sblevcmpkt{\sbradofcfm}{\sblevordofcfm}$ encodes that $\ofcfmean(\omega,\cdot)$ varies smoothly with the predictors for any $\omega \in \Omega$. This  enables approximation rates by controlling how complex the target surface is over $\domvX$; and it allows us to tie the depth and the width of the DFNN to the desired accuracy. 
In Assumption~\ref{ass:local_separation}, $\Delta(\rho)>0$ means that moving a distance $\rho$ from $\cfmean(\vx)$ raises the objective $\ofcfmean(\cfmean(\vx),\vx)$ uniformly in $\vx \in \R^{\dimX}$. 
This margin eliminates flat basins near the minimum and lets small objective errors imply small deviations away from the conditional \F mean. 
In addition, Assumption~\ref{ass:local_separation} implies the uniqueness of $\cfmean(\vx)$ for all $\vx\in\domvX$.  
Assumptions~\ref{ass:existence}--\ref{ass:local_separation} imply the existence of a DFNN such that the corresponding predictions are arbitrarily close to the conditional \F mean at an independent test point with arbitrarily high probability.

Additionally, under Assumptions~\ref{ass:cov_regularity} and \ref{ass:weight_regularity}, we show that there exists a DFNN such that the predictions are arbitrarily close to the conditional \F mean uniformly over all inputs $\vx \in \R^{\dimX}$. 
Assumption~\ref{ass:cov_regularity} imposes a uniform lower bound $\lambda_{\min}(\flyrCov)\ge \lambda_\star>0$ for all $\wtset_{\nlyr}\in\wtdom_{\nlyr}$, strengthening Assumption~\ref{ass:positive_definite_flyr_covariance} from pointwise positive definiteness to a global lower bound over the approximation class. 
Assumption~\ref{ass:weight_regularity} places norm constraints \citep{neys:15,bart:17} on $\prod_{l=1}^{\nlyr}\|\mwt_l\|_{\oprt}$ and $\max_{1\le l\le\nlyr}\|\vshift_l\|$, 
which uniformly control the magnitude of the final-layer representations which depends only on the depth $\nlyr$ and input dimension $\dimX$, but not on the particular weights---a key requirement for our approximation theorem. 

Given Assumptions~\ref{ass:existence}--\ref{ass:weight_regularity}, we establish Theorem~\ref{thm:approximation_theorem} below, which guarantees the universal approximation capability of DFNNs; 
we complement this theoretical result with empirical evaluations on both synthetic and real-world datasets in in Section~\ref{sec:empircal}.

\begin{theorem}
\label{thm:approximation_theorem}
\begin{enumerate}[label=(\roman*), wide,labelwidth=!,labelindent=0pt]
    \item\label{thm:approximation_theorem_sup} Under Assumptions \ref{ass:existence}--\ref{ass:weight_regularity}, for any $\eps > 0$, there exists $\delta =\delta_\eps>0$, $\nlyrapprox$ and $\wtsetapprox\in\wtdom_{\nlyrapprox}$ such that $\sup_{\vx  \in \domvX} d\{\predFNN(\vx;\wtsetapprox),\cfmean(\vx)\} < \eps$. 
    \item\label{thm:approximation_theorem_weak} Under Assumptions \ref{ass:existence}--\ref{ass:local_separation}, for any $\eps > 0$ and $\veps > 0$, there exists $\delta=\delta_{\eps,\veps}>0$, $\nlyrapprox$ and $\wtsetapprox$ such that $\prob(d\{\predFNN(\vX;\wtsetapprox),\cfmean(\vX)\} \geq \eps) < \veps$. 
\end{enumerate}
\end{theorem}

Next, we further investigate how the approximation error of the DFNN architecture translates into excess \F risk and prediction error. 
To this end, we require two other assumptions as follows.

\begin{assumption}
\label{ass: finite_second_moment}
For any $\nlyr\ge 1$, $\sup_{\vx \in \domvX,\,\wtset_{\nlyr}\in\wtdom_{\nlyr}} \E\{d^2(Y,\predFNN(\vx;\wtset_\nlyr))\} < \infty$.
\end{assumption}

\begin{assumption}\label{ass: curvature_pop}
There exists $\tilde\rho_\star$ such that for all $0 < \rho \leq \tilde\rho_\star$,
\bal \label{eq: loc_sep_risk}
& \Delta_\star (\rho) = \inf_{\f:\ \E[d\{\f(\vX),\cfmean(\vX)\}] \geq \rho} \{\risk(\f) - \risk(\cfmean)\} > 0. 
\eal
Additionally, there exists $\tilde \gamma_\star > 0$, 
$\tilde D > 0$ and $\tilde \alpha > 1$ such that for any predictor $\f(\cdot)$
\bal\label{eq: loc_curv_risk}
	& \risk(\f) - \risk(\cfmean) \geq  \left[\tilde D\E_{\vX}\{d(\f(\vX),\cfmean(\vX)) \}\right]^{\tilde\alpha} \quad \text{whenever}
    \\ & \E_{\vX}[d\{\f(\vX),\cfmean(\vX)\}] <\tilde \gamma_\star.
\eal
\end{assumption}

Assumption~\ref{ass: finite_second_moment} is a regularity assumption on the second moment of distance from the response, in particular, to the DFNN predictors with all possible architectures. 
Assumption~\ref{ass: curvature_pop} is a curvature condition that guarantees that the population \F objective function has sufficient curvature around its minimum $\ofcfmean(\cfmean(\vx),\vx)$, uniformly over $\vx \in \domvX$. 
Under Assumptions~\ref{ass: finite_second_moment}--\ref{ass: curvature_pop}, 
we establish in Corollary~\ref{cor: pred_approx} below that the excess risk of the population DFNN minimizer is of the order of the uniform approximation error $\Upsilon_\nlyr$, 
while its prediction error is of the order $\Upsilon_\nlyr^{1/\tilde\alpha}$, 
where 
\bal \label{eq: approx_error}
& \Upsilon_\nlyr = \inf_{\wtset_\nlyr \in \wtdom_\nlyr} \sup_{\vx \in \domvX} d\{\predFNN(\vx;\wtset_{\nlyr}),\cfmean(\vx)\} 
\quad \text{for } \nlyr\ge 1.
\eal 
Moreover, the universal approximation result in Theorem~\ref{thm:approximation_theorem} guarantees the existence of a sequence of network depths along which $\Upsilon_\nlyr$ converges to zero, and hence both the excess \F risk and prediction error decrease to zero along this sequence.

\begin{corollary} \label{cor: pred_approx}
For $\nlyr \geq 1$, with $\Upsilon_\nlyr$ defined in Eq.~\eqref{eq: approx_error}, we have under Assumptions~\ref{ass: finite_second_moment} and \ref{ass: curvature_pop}, there exists a universal constant $C >0$ such that,
\bal \label{eq: risk_approx} 
& \risk(\predFNN(\cdot;\trwtset_{\nlyr})) - \risk(\cfmean) \leq  C \Upsilon_\nlyr, \quad \text{and}
\\ & \E[d\{\predFNN(\vX;\trwtset_{\nlyr}),\cfmean(\vX)\}] \leq C \Upsilon_\nlyr^{1/\tilde\alpha},
\eal 
where the second statement in Eq.~\eqref{eq: risk_approx} holds for sufficiently small $\Upsilon_\nlyr$. Furthermore, under Assumptions \ref{ass:existence}--\ref{ass:weight_regularity}, there exists a sequence $\{\nlyr_k\}_{k \geq 1}$ such that $a_{\nlyr_k} \rightarrow 0$ as $k \rightarrow \infty$, and consequently,
\bal \nn
& \risk(\predFNN(\cdot;\trwtset_{\nlyr_k})) - \risk(\cfmean) \rightarrow 0, \quad \text{and}
\\ & \E[d\{\predFNN(\vX;\trwtset_{\nlyr_k}),\cfmean(\vX)\}] \rightarrow 0
\eal 
as $k \rightarrow \infty$.
\end{corollary}

Next, we investigate the generalization properties of DFNN and derive corresponding risk bounds. As a first step, Theorem~\ref{thm: output_layer_convergence} below establishes the consistency and rate of convergence of the empirical predictor output of a trained DFNN to its population counterpart, uniformly over the neural network parameter space. 
This result provides a key ingredient for analyzing the generalization behavior of the proposed DFNN estimator. 
To state Theorem~\ref{thm: output_layer_convergence}, we first introduce notations and the assumptions required to establish it. Define 
\bal \label{eq: output_layer_set}
\widetilde\Omega_1 &= \{ \cfmean(\vx): \vx \in \domvX\}, 
\\ \widetilde\Omega_2 &= \{\predFNN(\vx;\wtset_\nlyr): \vx \in \domvX, \wtset_\nlyr \in \wtdom_\nlyr\}, \quad \text{and}
\\ \widetilde\Omega &= \{ \cfmean(\vx): \vx \in \domvX\} \cup \{\predFNN(\vx;\wtset_\nlyr): \vx \in \domvX, \wtset_\nlyr \in \wtdom_\nlyr\}.
\eal
Here, $\widetilde\Omega_1$ denotes the set of population predictor outputs, while $\widetilde\Omega_2$ denotes the set of DFNN predictor outputs over the input and neural network parameter spaces. %In Lemma~\ref{lma : totally_bounded_output_layer} in Section~\ref{sec:aux_lemma} of the \sppl, 
We establish that $\widetilde\Omega$ is totally bounded under Assumptions~\ref{ass:existence}--\ref{ass: finite_second_moment} and \ref{ass:local_separation_ofFNN}--\ref{ass:weight_frobenius} that is needed in establishing Theorem~\ref{thm: output_layer_convergence}.

\begin{assumption}
\label{ass:local_separation_ofFNN}
There exists $\radofpredFNN>0$ such that for any $\eta \in(0,\radofpredFNN]$ and any integer $\nlyr\ge 1$, 
\bal\nn
   \tilde{\Delta}_\nlyr(\eta)=\inf_{\wtset_\nlyr \in \wtdom_\nlyr} \inf_{\vx \in \domvX}\inf_{d\{\omega,\predFNN(\vx;\wtset_\nlyr)\}\geq \eta} \{ \ofFNN(\omega,\vx;\wtset_\nlyr) - \ofFNN(\predFNN(\vx;\wtset_\nlyr),\vx;\wtset_\nlyr)\} > 0.
\eal 
\end{assumption}

\begin{assumption}
\label{ass:weight_frobenius}
There exist constants $\uppwtfrob_{l}>0, \ l=1, \dots, \nlyr$ such that $\wtset_{\nlyr}$ satisfies $ \|\tilde\mwt_{l}\|_\frob \leq \uppwtfrob_{l}$ for any integer $\nlyr\ge 1$, where $\tilde\mwt_{l}$ is given by
\bal \nn 
& \tilde\mwt_{l} =
\begin{pmatrix}
\mwt_{l} & \vshift_l\\
\mathbf 0^\top & 1
\end{pmatrix}.
\eal
\end{assumption}

\begin{assumption}\label{ass: predFNNhat_tight}
For every $r>0$, there exists a totally bounded set
$K_{r}\subset\Omega$ such that
\bal \nn
\liminf_{n\to\infty}
\prob\left(
\predFNNhat(\vx;\wtset_\nlyr) \in K_r
\text{ for all } \vx \in \domvX,\ \wtset_\nlyr\in\wtdom_\nlyr
\right)
\ge 1-r.
\eal
\end{assumption}

\begin{assumption}\label{ass: curvature}
There exists $\gamma_\star >0$, 
$D > 0$ and $\alpha > 1$ such that
\bal\nn
\inf_{\vx \in \domvX} \inf_{\wtset_\nlyr \in \wtdom_\nlyr} \inf_{d(\omega,\predFNN(\vx;\wtset_\nlyr)) < \gamma_\star} \big\{\ofFNN(\omega,\vx;\wtset_\nlyr)-\ofFNN(\predFNN(\vx;\wtset_\nlyr),\vx;\wtset_\nlyr)&
\\ -Dd^\alpha(\omega,\predFNN(\vx;\wtset_\nlyr)) \big\}& \geq 0.
\eal
\end{assumption}

\begin{assumption} \label{ass: entropy}
For any $\delta > 0$, define the entropy integrals 
\bal \label{eq: entropy_integral}
& \tilde J_\delta = \int_0^{\delta} \big\lbrace  \log N\big( u, \widetilde\Omega,d\big)\big\rbrace^{1/2} \diffop u \quad \text{and}\quad 
J_\delta = \int_0^{\delta} \sup_{\omega \in \widetilde\Omega} \big\lbrace  \log N\big( u, \ballatomega_{\delta},d\big)\big\rbrace^{1/2} \diffop u.
\eal
There exists $\kappa > 0$ such that $\tilde J_\delta = O(\delta^{\kappa})$ and $J_\delta = O(\delta^{\kappa})$ as $\delta \rightarrow 0$.
\end{assumption}

Assumption~\ref{ass:local_separation_ofFNN} is a standard assumption for $M$-estimators. Similar to Assumption~\ref{ass:local_separation}, it implies $\predFNN(\vx;\wtset_{\nlyr})$ is the unique minimizer in Eq.~\eqref{eq:dfnn_output_layer} for any $\vx\in\domvX$ and $\wtset_{\nlyr}\in\wtdom_{\nlyr}$ and guarantees that small deviations from the minimum of the objective function $\ofFNN(\cdot,\vx;\wtset_{\nlyr})$ lead to small deviations from $\predFNN(\vx;\wtset_{\nlyr})$. 
Assumption~\ref{ass:weight_frobenius}, commonly imposed in the neural network literature \citep{golo:20}, is needed to establish the consistency and rate of convergence of the proposed DFNN empirical objective to its population counterpart uniformly over the weight class $\wtdom_\nlyr$. This uniform convergence is essential for proving the convergence of the DFNN output layer to its population counterpart, uniformly in the weights, which is subsequently needed for the generalization analysis. 
Alternative approaches would be to control the metric entropy of the neural network function class directly, as in \citet{bart:17}. 
However, the analysis of \citet{golo:20} based on Frobenius norm constraints yields complexity bounds that avoid an explicit dependence on the potentially large number of network parameters, and can therefore provide sharper rates when the input dimension $\dimX$ is moderate relative to the size of the network; see also \cite{sell:24} for more recent depth-independent bounds. Assumption~\ref{ass: predFNNhat_tight} guarantees the uniform asymptotic tightness of the family $\{\predFNNhat(\vx;\wtset_\nlyr): \vx \in \domvX,\ \wtset_\nlyr\in\wtdom_\nlyr\}$, 
which is commonly imposed in establishing uniform consistency of such $M$-estimators. 
In particular, Assumption~\ref{ass: predFNNhat_tight} allows us to restrict attention to a totally bounded subset of $\msp$ on which $\sup_{\vx\in\domvX,\, \wtset_{\nlyr}\in\wtdom_{\nlyr}} |\ofFNNhat(\cdot,\vx;\wtset_\nlyr) -\ofFNN(\cdot,\vx;\wtset_\nlyr)|$ is uniformly controlled. 
Consequently, uniform convergence of $\ofFNNhat(\omega,\vx;\wtset_{\nlyr})$ to $\ofFNN(\omega,\vx;\wtset_{\nlyr})$ over $\omega\in\msp$, $\vx\in\domvX$, and $\wtset_{\nlyr}\in\wtdom_{\nlyr}$ can be reduced to establishing the corresponding convergence over a totally bounded subset of $\msp$, uniformly in $\vx\in\domvX$ and $\wtset_{\nlyr}\in\wtdom_{\nlyr}$.  Assumption~\ref{ass: curvature} is a uniform curvature condition on the objective function $\ofFNN(\cdot,\vx;\wtset_{\nlyr})$ near its  minimizer $\predFNN(\vx;\wtset_\nlyr)$, which is needed to establish the rate of convergence of $\predFNNhat(\vx;\wtset_\nlyr)$ to $\predFNN(\vx;\wtset_\nlyr)$. 
Assumption~\ref{ass: entropy} is a  standard entropy condition to control the behavior of the empirical process $\frac{1}{n} \sum_{i=1}^n {\gfrwt}\{\funlyr_{\nlyr}(\vx;\wtset_{\nlyr}),\funlyr_{\nlyr}(\vX_i;\wtset_{\nlyr})\} d^2(Y_i,\omega)-\ofFNN(\omega,\vx;\wtset_{\nlyr})$ near $\predFNN(\vx;\wtset_{\nlyr})$ and is satisfied by a broad rance of metric spaces. 
In particular, Assumption~\ref{ass: entropy} holds for any space $(\msp,d)$ such that $\log N(\eps,\msp,d) = O(\eps^{-\varsigma})$ with some $\varsigma<2$. 
Examples include bounded subsets of finite-dimensional Euclidean spaces, 
compact finite-dimensional Riemannian manifolds, 
uniformly bounded VC classes of functions, 
the space of distributions on a compact interval under Wasserstein metric, 
the space of networks with a fixed number of nodes and bounded edge weights represented by graph
Laplacians or adjacency matrices, 
the space of symmetric positive definite matrices with a fixed dimension and uniformly bounded eigenvalues under the Frobenius, log-Euclidean, or affine-invariant metric, 
and Billera--Holmes--Vogtmann (BHV) space \citep{bill:01} of rooted metric phylogenetic trees with a fixed number of tips and uniformly bounded branch lengths 
\citep[see][for related discussions]{dube:24,chen:26:2}.

\begin{theorem}
\label{thm: output_layer_convergence}
For any integer $\nlyr\ge 1$, for a  sufficiently large constant $C > 0$, under Assumptions~\ref{ass:existence}--\ref{ass: finite_second_moment} and 
\ref{ass:local_separation_ofFNN}--\ref{ass: predFNNhat_tight}, we have
\bal \label{eq: output_layer_uniform_consistency}
  \sup_{\wtset_\nlyr \in \wtdom_\nlyr} \sup_{\vx  \in \domvX}  d\{\predFNNhat(\vx;\wtset_\nlyr),\predFNN(\vx;\wtset_\nlyr)\}  = \op\left(1\right)
\eal
as $n \rightarrow \infty$. Furthermore, under Assumptions~\ref{ass:existence}--\ref{ass: finite_second_moment} and \ref{ass:local_separation_ofFNN}--\ref{ass: entropy}, 
\bal \label{eq: output_layer_rate}
  n^{\zeta} \sup_{\wtset_\nlyr \in \wtdom_\nlyr} \sup_{\vx  \in \domvX}  d\{\predFNNhat(\vx;\wtset_\nlyr),\predFNN(\vx;\wtset_\nlyr)\}  = \Op\left(1\right),
\eal
where $\zeta=[2\{\alpha-\min(\kappa,1)\}]\inv$.
\end{theorem}

We next study the statistical performance of the proposed DFNN estimator obtained using the empirical \F risk minimizer, 
$\spredY(\vx) 
= \predFNNhat(\vx;\trwtsethat_{\nlyr})$ for $\vx\in\R^{\dimX}$ as defined in Eq.~\eqref{eq:dfnn_prediction_emp}. 
To formalize this, let $\mathcal{M} = \left\lbrace \f\colon\R^\dimX \to \Omega \mid \risk(\f) < \infty \right\rbrace$ denote the class of all predictors with finite \F risk, 
where $\risk(\f)$ is the \F risk of predictor $\f$, as defined in Eq.~\eqref{eq:risk}. 
Observe that the regression target, the conditional \F mean $\cfmean$  belongs to $\mathcal{M}$, and satisfies
\bal \nn
\cfmean \in \argmin_{\f \in \mathcal{M}} \risk(\f).
\eal
Hence, $\cfmean$ is a global minimizer of the \F risk among all predictors in $\mathcal M$. 
In Theorem~\ref{thm: risk_and_pred_error_bound}, we compare the \F risk of the proposed DFNN estimator $\risk\{\predFNNhat(\cdot;\trwtsethat_{\nlyr})\}$ to the optimal \F risk $\risk(\cfmean)$ attained by the conditional \F mean $\cfmean$ over the class $\mathcal M$. 
Furthermore, we establish the order in probability of the prediction error $\E_{\vX}[d\{\predFNNhat(\vX;\trwtsethat_{\nlyr}),\cfmean(\vX)\}]$. 

\begin{assumption}\label{ass: curvature_for_risk}
There exists $D_\star > 0$ and $\alpha_\star > 1$ such that for any $\wtset_\nlyr, \wtset'_\nlyr \in \wtdom_\nlyr$, 
\bal\nn
&\ofFNN(\predFNN(\vX;\wtset'_{\nlyr}),\vX;\wtset_\nlyr)-\ofFNN(\predFNN(\vX;\wtset_\nlyr),\vX;\wtset_\nlyr) 
\\ \geq\, & D_\star d^{\alpha_\star}(\predFNN(\vX;\wtset'_{\nlyr}),\predFNN(\vX;\wtset_\nlyr)) \}, 
\quad\text{almost surely.}
\eal
\end{assumption}

\begin{theorem} \label{thm: risk_and_pred_error_bound}
Under Assumptions~\ref{ass:existence}--\ref{ass: finite_second_moment} and \ref{ass:local_separation_ofFNN}--\ref{ass: curvature_for_risk}, 
\bal \label{eq: risk_bound}
& \risk\{\predFNNhat(\cdot;\trwtsethat_{\nlyr})\} - \risk(\cfmean) = \Op\left( \Gamma^\star_{n,\nlyr}+\Upsilon_\nlyr\right)
\eal
where $\Upsilon_\nlyr$ as defined in Eq.~\eqref{eq: approx_error}, and 
\bal \nn
\Gamma^\star_{n,\nlyr} 
= \delta^{1/(2\alpha_\star)}_{n,\nlyr} 
+ \frac{1}{\sqrt{n}} \bigg[\E\bigg\lbrace \Big(\int_{C \sqrt{\delta_{n,\nlyr}}}^{C} u^{1/\alpha_\star - 1}\sqrt{ \log N\big(u, \mathcal{G}, L_2(P_n)\big)} \diffop u\Big)^2  \bigg\rbrace\bigg]^{1/2} 
+ n^{-\zeta},
\eal
for a universal constant $C> 0$. 
Here, 
\bal \label{eq: delta_nL}
& \delta_{n,\nlyr} = c \uppwtop \left\lbrace \sqrt{\dimX} + \nlyr\uppshift \right \rbrace \wlyr_\nlyr  \left( \prod_{l=1}^\nlyr \uppwtfrob_l \right) \sqrt{\frac{\dimX\nlyr}{n}},
\eal
for some constant $c>0$, 
$\zeta$ is as defined in Theorem~\ref{thm: output_layer_convergence}, 
and $N\left(u, \mathcal{G}, L_2(P_n)\right)$ denotes the covering number of the function class $\mathcal G=\{ \funlyr_{\nlyr}(\cdot;\wtset_{\nlyr}): \wtset_\nlyr \in \wtdom_\nlyr\}$ 
with $L_2(P_n)$-balls of radius $u$  for any $u > 0$. 
Additionally, under Assumptions~\ref{ass:existence}--\ref{ass: curvature_for_risk},
\bal \label{eq: pred_bound}
\E_{\vX}\left[d\{\predFNNhat(\vX;\trwtsethat_{\nlyr}),\cfmean(\vX)\}\right] = \Op\left( (\Gamma^\star_{n,\nlyr}+\Upsilon_\nlyr)^{1/\tilde \alpha}\right).
\eal
Furthermore, under Assumptions \ref{ass:existence}--\ref{ass: finite_second_moment}, there exists a sequence $\{\nlyr_k\}_{k \geq 1}$ such that $\Upsilon_{\nlyr_k} \rightarrow 0$ as $k \rightarrow \infty$, and consequently,
\bal \nn
&  \risk\{\predFNNhat(\cdot;\trwtsethat_{\nlyr_k})\} - \risk(\cfmean) = \Op\left( \Gamma^\star_{n,\nlyr}\right), \quad \text{and}
\\ & \E_{\vX}\left[d\{\predFNNhat(\vX;\trwtsethat_{\nlyr_k}),\cfmean(\vX)\}\right] = \Op\left( (\Gamma^\star_{n,\nlyr})^{1/\tilde \alpha}\right)
\eal 
as $k \rightarrow \infty$ and $n \rightarrow \infty$.
\end{theorem}

\begin{remark}\label{rem: entropy_rate}
The entropy term in Theorem~\ref{thm: risk_and_pred_error_bound} can be made explicit using existing covering-number bounds for deep neural networks. 
In particular, we prove by using Theorem~3.3 of \citet{bart:17} that 
$\log N\left(u,\mathcal{G},L_2(P_n)\right) \le C'(nu^2)\inv$ for some constant $C'>0$, 
and hence $\Gamma^\star_{n,\nlyr} = O(n^{-1/(4\alpha_\star)}
+n^{-\zeta})$ for any fixed network architecture.
\end{remark}

\begin{remark} \label{rem: entropy_rate_fast}
The norm-based covering bounds discussed in Remark~\ref{rem: entropy_rate} accommodate large fixed network architectures through norm control on the parameters, i.e., the size of the architecture enters through the finite constant $C_{\nlyr}$, rather than through a direct dependence on the total number of network parameters. In contrast, for architectures where the total number of parameters is manageable, for example sparse neural network architectures, a complementary faster rate can be obtained from the classical parameter-count-based results in \citet{anth:09}. With $P_{\nlyr}
=
\sum_{l=1}^{\nlyr}
\wlyr_l(\wlyr_{l-1}+1),
\
\wlyr_0=\dimX,$ denoting the total number of parameters in the hidden layers, it can be derived that 
\bal \nn 
\log N\left( u,\mathcal G,L_\infty\right)
\leq
C P_{\nlyr}
\log(C/u),
\eal
for some $C > 0$.  
Since $\int_0^C u^{1/\alpha_\star-1}
\{\log(C/u)\}^{1/2}\,du
<\infty,$ 
a modification of Theorem~\ref{thm: risk_and_pred_error_bound} in this regime, adapted to $L_\infty$-covering bounds for $\mathcal G$, gives  
\bal \nn
& \risk\{\predFNNhat(\cdot;\trwtsethat_{\nlyr})\} - \risk(\cfmean)
=
\Op\left(
\sqrt{
\frac{
P_{\nlyr}
}{n}
}
+
n^{-\zeta}
+
\Upsilon_{\nlyr}
\right).
\eal
Under Assumption~\ref{ass: curvature_pop}, the corresponding prediction error rates follow similarly to Theorem~\ref{thm: risk_and_pred_error_bound} by raising the respective excess-risk rates to the power $1/\tilde\alpha$.
\end{remark}

Finally, in Remark~\ref{rem: optimization_guarantee} we clarify the scope of our theoretical results.

\begin{remark} \label{rem: optimization_guarantee}
The theoretical analysis in this paper focuses on the approximation and statistical properties of DFNNs, analogous to corresponding results in \citep{bart:17,schm:20,kohl:21}, rather than the algorithmic aspects of training. In particular, we do not study the theoretical properties of optimization algorithms such as stochastic gradient descent (SGD), the geometry of the empirical \F risk landscape, including the presence of saddle points or local minima, or guarantees for attaining exact or approximate empirical risk minimizers. Throughout, we assume access to optimization procedures that minimize the empirical \F risk in Eq.~\eqref{eq:dfnn_risk_emp}, and study the statistical properties and generalization performance of the resulting estimator. Bridging the statistical theory developed here with an analysis of the optimization dynamics of DFNN training, including guarantees for approximate empirical risk minimization and their resulting statistical and prediction errors, is an important direction for future research.
\end{remark}
    
\section{Experiments}\label{sec:empircal}

\subsection{Implementation Details}\label{sec:impl_details}

We implement deep neural networks in \texttt{PyTorch} with default parameter initializations for linear layers. 
Training uses stochastic gradient descent (SGD) with a fixed momentum of $0.9$. 
For each experiment, we select the architectural and regularization-related hyperparameters---the number of hidden layers $\nlyr$, the width of hidden layers $\{\wlyr_{l}\}_{l=1}^{\nlyr}$, which are taken as the same value across first $(\nlyr-1)$ hidden layers, i.e., $\wlyr_{l}=\wlyr_{1}$ for all $l=1,\dots,\nlyr-1$, the learning rate (LR), and the dropout rate (DR)---via a grid search, 
and then use the corresponding choices for that experiment. 
Grids of candidate values and the selected values are listed in Section~\ref{sec:arch_rgl} in the \sppl.

To mitigate overfitting for the proposed DFNN, we implement a validation-based early stopping rule \citep[Algorithm~7.1 in Chapter~7.8 of][]{good:16}. 
The original training data is randomly split into a training set and a validation set in a $4$-to-$1$ proportion. 
The DFNN is trained only on the training set. 
At each epoch, we evaluate the validation loss, given by the mean squared prediction error (MSPE) on the validation set.
After a fixed burn-in period of $\nepint$ epochs, 
we stop training if the validation loss at the current epoch fails to improve by at least $\tolr$ below the lowest validation loss observed so far 
over $\npatn$ 
consecutive epochs. 
Then we use the weights with which the DFNN achieves the lowest validation loss observed before stopping. 
The selection of the hyperparameters in the early stopping rule, i.e., $\nepint$, $\tolr$, and $\npatn$, are detailed in Section~\ref{sec:earlystop} in the \sppl for each scenario in the experiments. 

\subsection{Experiment Setups}\label{sec:expr_setups}

To evaluate the effectiveness of our proposed DFNN framework, we perform empirical experiments on both synthetic data and real-world data. 
Our experiments cover responses lying in three different types of metric spaces. 
First, we consider the Wasserstein space of univariate probability measures endowed with the $2$-Wasserstein metric \citep{vill:08}. 
We simulate data with responses in the Wasserstein space in  Experiment~\ref{case:distn_lowdim2} below. 
Second, we consider the space of graph Laplacian matrices with the Frobenius metric and simulate data with network-valued responses in this space in Experiment~\ref{case:network} below. 
Third, we handle U.S.\ states' employment occupational compositions as the responses in Experiment~\ref{case:employment_composition}, which are viewed as random objects lying in the $8$-simplex $\Delta^8 = \{\va\in[0,1]^9: \sum_{j=1}^{9}a_{j} = 1\}$, endowed with the Aitchison distance given by the Euclidean metric between the centered log-ratios of compositions \citep{aitc:86}. 
Fourth, we investigate the U.S.\ states' electricity generation data where electricity generation source compositions are mapped to the positive orthant $\sphere^{11}_+ = \{\vu\in [0,1]^{12}:\|\vu\|=1\}$
of the unit sphere $\sphere^{11}=\{\vu\in\R^{12}:\|\vu\|=1\}$ by the square root transformation \citep{scea:11}, which is endowed with the geodesic distance on the sphere. 

In each experiment with simulated data (Experiments~\ref{case:distn_lowdim2}--\ref{case:network}), we compare the proposed DFNN with other methods, including: global Fr\'echet regression (GFR) \citep{pete:19}, local Fr\'echet regression with sufficient dimension reduction (SDR) \citep{zhan:24:2}, and deep Fr\'echet regression (DFR) \citep{iao:25}, and end-to-end deep learning for metric space-valued outputs (E2M) \citep{zhou:26:2}. 
For DFR, we follow the selection of hyperparameters for the architecture and regularization in \cite{iao:25}; see Table~\ref{tab:DFR-hp} in Section~\ref{sec:impl_others} in the \sppl for details. 
For our DFNN, we choose the hyperparameters for the architecture and regularization following the procedures in Section~\ref{sec:impl_details}, 
and the grids of candidate values and chosen values are listed in Section~\ref{sec:arch_rgl} in the \sppl. 
We report results for four variants of DFNN, obtained by crossing the choices of the size of the network architecture and the treatment of the training loss. 
Regarding the architecture size,  
the ``base'' variants (\dfnnbase and \adfnnbase) use a smaller network architecture matching that of DFR, both to enable a fair comparison with DFR and to demonstrate that DFNN performs well even under this more restricted architectural setting. 
Regarding the training loss, 
for distributional and network-valued responses $\{Y_i\}_{i=1}^{n}$, 
since the weights $\gfrwthat\{\funlyr_{\nlyr}(\vx;\wtset_{\nlyr}),\funlyr_{\nlyr}(\vx';\wtset_{\nlyr})\}$ in Eq.~\eqref{eq:dfnn_output_weight_emp} can be negative, 
the weighted average $\frac{1}{n} \sum_{i=1}^n {\gfrwthat}\{\funlyr_{\nlyr}(\vx;\wtset_{\nlyr}),\funlyr_{\nlyr}(\vX_i;\wtset_{\nlyr})\} Y_i$ may not be a valid distribution or graph Laplacian and hence need not to be the minimizer $\predFNNhat(\vx;\wtset_{\nlyr})$ in Eq.~\eqref{eq:dfnn_output_layer_emp}. 
This makes computing the gradients of the exact training loss, i.e., the empirical \F risk 
$\efrisk\{\predFNNhat(\cdot;\wtset_{\nlyr})\}$ in Eq.~\eqref{eq:dfnn_risk_emp}, more involving, which needs to be handled case-by-case for each metric space (see Section~\ref{sec:gradient_comp} for details). 
\dfnn and \dfnnbase execute with the training loss given by the exact empirical \F risk 
$\efrisk\{\predFNNhat(\cdot;\wtset_{\nlyr})\}$ in Eq.~\eqref{eq:dfnn_risk_emp}, at the
price of gradient computation being derived separately for each metric space. 
In contrast, \adfnn and \adfnnbase instead train with an approximate training loss that uses $\frac{1}{n} \sum_{i=1}^n {\gfrwthat}\{\funlyr_{\nlyr}(\vx;\wtset_{\nlyr}),\funlyr_{\nlyr}(\vX_i;\wtset_{\nlyr})\} Y_i$ as the surrogate of the minimizer $\predFNNhat(\vx;\wtset_{\nlyr})$ in Eq.~\eqref{eq:dfnn_output_layer_emp}, so that training requires only standard automatic differentiation. 
We note that this approximation is only applied to the training loss, 
and that the final predictions $\spredY(\vx) = \predFNNhat(\vx;\trwtsethat_{\nlyr})$ in Eq.~\eqref{eq:dfnn_prediction_emp} are computed exactly 
so that they always lie in the corresponding metric space. 

For the real-world applications (Experiments~\ref{case:employment_composition}--\ref{case:energy_sphere}), the set of competing methods varies depending on the availability of implementations for the corresponding metric spaces. 
We therefore compare DFNN with GFR, SDR, DFR and E2M in
Experiment~\ref{case:employment_composition}, and with GFR and SDR in
Experiment~\ref{case:energy_sphere}, since DFR and E2M are unavailable for
sphere-valued responses \citep{iao:25,zhou:26:2}. 

We evaluate the performance of each method by the mean squared prediction error (MSPE). 
For each experiment with simulated data (Experiments~\ref{case:distn_lowdim2}--\ref{case:network}), for each Monte Carlo run out of $\nMC = 250$ runs, we generate a training sample of size $n$, $\{(\vX_{i},Y_{i})\}_{i=1}^{n}$, and a test set of 100 additional pairs $\{(\vX_{n+i},Y_{n+i})\}_{i=1}^{100}$, 
and evaluate the MSPE on the test set, which is defined as  
\bal \label{eq:MSPE_simu}
\mathrm{MSPE} = \frac{1}{100} \sum_{i=1}^{100} d^2 (\spredY_{n+i}, Y_{n+i}),
\eal
where $\spredY_{n+i}$ is the prediction of $Y_{n+i}$ given by the corresponding method, e.g., for the proposed DFNN, $\spredY_{n+i}=\spredY(\vX_{n+i})$ with $\spredY(\cdot)$ defined in Eq.~\eqref{eq:dfnn_prediction_emp}.
For the experiments applying to real-world data (Experiments~\ref{case:employment_composition} and~\ref{case:energy_sphere}), we perform $10$-fold cross-validation. The sample is randomly split into $10$ folds, and each fold is left out as a test set, respectively. The prediction of a response $Y_i$ that belongs to the $k$th fold for some $k=1,\dots,10$ is given by the model trained on the other $9$ folds except for the $k$th fold, denoted by $\spredY_{i,\mathrm{cv}}$, for $i=1,\dots,n$.
The MSPE is defined as 
\bal \label{eq:MSPE_real}
\mathrm{MSPE} = \frac{1}{n} \sum_{i=1}^{n} d^2 (\spredY_{i,\mathrm{cv}}, Y_{i}),
\eal 
We repeat the random splits $100$ times and each time evaluate the MSPE as defined in Eq.~\eqref{eq:MSPE_real}. 

In the following, we describe the data in each experiment. 

\begin{case}\label{case:distn_lowdim2}
(Simulated data with distributional responses.)
Here we investigate performance of predicting distributional responses. In each Monte Carlo run, a training sample of $n$ i.i.d. pairs, $\{(\vX_i,Y_i)\}_{i=1}^{n}$, are generated as follows. 
For each $i=1,\dots,n$,  we generate the predictors $\vX_i=(X_{i,1}\dots,X_{i,10})\tps$ such that $(X_{i,1},\dots,X_{i,4})\tps$, $X_{i,5},\dots,X_{i,10}$ are mutually independent and that 
\bgt \nn
(X_{i,1}, \dots, X_{i,4})\tps \sim \normal(\bm{0}, \mathbf{\Sigma}_{0.1}), \quad \mathbf{\Sigma}_{\rho} = 
\begin{pmatrix}
1 & \rho & \cdots & \rho \\
\rho & 1 &  \cdots & \rho\\
\vdots & \vdots &  \ddots & \vdots \\
\rho & \rho &  \cdots & 1
\end{pmatrix}\in\R^{4 \times 4},\\
X_{i,j} \sim \normal(1,1), \quad j \in \{5,6,7,8,9\},\\
X_{i,j} \sim \bernoulli(0.3), \quad j \in \{10\}.
\egt
For each $i=1,\dots,n$, define 
\bal\nn
\mu_i &= X_{i,1}\cos(\pi X_{i,2}) - 0.5 X_{i,3}^2 + 4\log(1+X^2_{i,4}) - 4(1+|X_{i,5}|)\inv, 
\\ \theta_i &= 0.5+3.5\expit\{2X_{i,6}X_{i,10} + 2 X_{i,9}^2\sin(\pi X_{i,8})
+ 4(1+X_{i,7})\inv\},
\eal
where $\expit(t)=1/(1+e^{-t})$ for $t\in\R$. 
For each $i=1,\dots,n$, conditioning on $\vX_i$, we generate $\eta_{i}$ and $\sigma_{i}$ as 
\[
\eta_{i}\mid \vX_i \sim \normal(\mu_{i},0.5^2)
\quad \text{and}\quad
\sigma_{i}\mid \vX_i \sim \Gammadistn(\theta_{i}^2,\theta_{i}^{-1})
\]
and define $Y_i = \normal(\eta_i,\sigma_i^2)$. 
Furthermore, for each $i=1,\dots,n$, we draw a random sample of independent data points $\{Z_{i,k}\}_{k=1}^{100}$ from $Y_i$, 
and use the empirical quantile function of $\{Z_{i,k}\}_{k=1}^{100}$ as the surrogate of $Y_i$ in the training data. 
We perform the experiment with training samples of size $n\in\{200,500,1000\}$, respectively. 
\end{case}

\begin{case}\label{case:distn_highdim} 
(Simulated data with distributional responses and potentially high-dimensional predictors.)
Here we investigate performance of predicting distributional responses from potentially high-dimensional predictors. Let $\dimX=100$. In each Monte Carlo run, a training sample of $n$ i.i.d. pairs, $\{(\vX_i,Y_i)\}_{i=1}^{n}$, are generated as follows. 
For each $i=1,\dots,n$,  we generate the predictors $\vX_i=(X_{i,1}\dots,X_{i,\dimX})\tps$ such that $(X_{i,1},\dots,X_{i,\frac{\dimX}{10}})\tps$, $X_{i,\frac{\dimX}{10}+1},\dots,X_{i,\dimX}$ are mutually independent and that 
\bgt \nn
(X_{i,1}, \dots, X_{i,\frac{9}{10}\dimX})\tps \sim \normal\{\bm{0}, \diag(\mathbf{\Sigma}_{0.2},\mathbf{\Sigma}_{0.5})\}, \quad \mathbf{\Sigma}_{\rho} = 
\begin{pmatrix}
1 & \rho & \cdots & \rho \\
\rho & 1 &  \cdots & \rho\\
\vdots & \vdots &  \ddots & \vdots \\
\rho & \rho &  \cdots & 1
\end{pmatrix}\in\R^{(\frac{9}{20}\dimX) \times (\frac{9}{20}\dimX)},\\
X_{i,j} \sim \normal(1,1), \quad j \in \{\tfrac{9}{10}\dimX+1,\dots,\dimX-3\}, \\
X_{i,j} \sim \bernoulli(0.3 + 0.1(\dimX-j)), \quad j \in \{\dimX-2,\dimX-1,\dimX\}. 
\egt
For each $i=1,\dots,n$, define 
\bal\nn
\theta_{i,0} &= X_{i,1}\cos(\pi X_{i,2}) - 0.5 X_{i,3}^2 + 4\log(1+X^2_{i,4}) - 4(1+|X_{i,5}|)\inv, 
\eal
where $\expit(t)=1/(1+e^{-t})$ for $t\in\R$, and  
for each $j=1,\dots,20$, define 
\bal\nn
\theta_{i,j} &= 2|X_{i,5(j-1)+1}X_{i,5(j-1)+2}| + 2 X_{i,5(j-1)+3}^2\sin^2(\pi X_{i,5(j-1)+4})\\
&\quad + 4(1+|X_{i,5(j-1)+5}|)\inv;
\eal
then conditioning on $\vX_i$, we generate $q_{i,0}$ and $\Delta q_{i,j}$ for $j=1,\dots,20$ as 
\bal\nn
q_{i,0} \mid \vX_i\sim\normal(\theta_{i,0},0.5^2)
\quad\text{and}\quad
\Delta q_{i,j} \mid \vX_i\sim\Gammadistn(\theta_{i,j}^2,\theta_{i,j}\inv)
\eal
and define 
\bal\nn
q_{i,j} &= q_{i,0} + \sum_{k=1}^{j}\Delta q_{i,k}. 
\eal
Then for each $i=1,\dots,n$, we define $Y_i$ as the distribution with $(j/20)$-th quantile given by $q_{i,j}$ for each $j=0,1,\dots,20$ and $[u(j-1)/20 + (1-u)j/20]$-th quantile given by $uq_{i,j-1}+(1-u)q_{i,j}$ for $u\in(0,1)$ and $j=1,\dots,20$. 
Furthermore, for each $i=1,\dots,n$, we draw a random sample of independent data points $\{U_{i,k}\}_{k=1}^{100}$ from $\unif(0,1)$ and define $Z_{i,k} = F\inv_{Y_i}(U_{i,k})$ for $k=1,\dots,100$, where $F\inv_{Y_i}$ is the quantile function of $Y_i$---as such, $\{Z_{i,k}\}_{k=1}^{100}$ is a sample of independent data points drawn from $Y_i$. 
We use the empirical quantile function of $\{Z_{i,k}\}_{k=1}^{100}$ as the surrogate of $Y_i$ for each $i=1,\dots,n$ in the training data. 
We perform the experiment with training samples of size $n\in\{100,200,500,1000\}$, respectively. 
\end{case}

\begin{case}\label{case:network}
(Simulated data with network-valued responses represented by graph Laplacian matrices.) 
We consider responses being graph Laplacian matrices of networks with $\nnode$ nodes. 
For each Monte Carlo run, we first generate a binary mask matrix $\mathbf{\mask}\in\{0,1\}^{\nnode \times \nnode}$ encoding the presence of edges, with $\mask_{k,\ell}=\mask_{\ell ,k}\sim\bernoulli(0.3)$ for any $k,\ell\in\{1,2,\dots,\nnode\}$ with $k<\ell$ independently, and $\mask_{k,k}=0$ for all $k\in\{1,2,\dots,\nnode\}$. 
For each $i=1,\dots,n$, we generate $X_{i,1},\dots,X_{i,10}$ i.i.d. from $\unif(0,1)$, and define $\vX_i=(X_{i,1},\dots,X_{i,10})\tps$. 
Conditioning on $\vX_i$ and $\mathbf{\mask}$, we generate a symmetric edge-weight matrix $\mathbf{\edgewt}_i = (\edgewt_{i,k,\ell})_{1\le k,\ell\le \nnode}$ on the connected edges as:
\bal\nn 
\edgewt_{i,k,\ell}
=\begin{cases}
\sin\!\Big(\frac{(k+\ell)\pi}{2\nnode}\Big)\cdot\frac{1}{\,|X_{ik}|+1\,}\cdot\big(2+X_{i\ell}^2\big)
+\varepsilon_{i,k,\ell} &\text{if } k<\ell \text{ and } \mask_{k,\ell}\ne 0\\
\edgewt_{i,\ell,k} &\text{if } k>\ell \text{ and } \mask_{k,\ell}\ne 0\\
0 &\text{otherwise}
\end{cases} 
\eal
where $\varepsilon_{i,k,\ell}\sim\unif(-a,a)$  are additional noises generated independently across $(i,k,\ell)$ with $a=0$ corresponding to noiseless scenarios and larger values of $a$ corresponding to higher levels of noise. 
Let $\mathbf{D}_i$ be the degree matrix of $\mathbf{\edgewt}_i$, i.e., a diagonal matrix with diagonal elements given by $\mathbf{\edgewt}_i \bm{1}_{\nnode}$, where $\bm{1}_{\nnode}$ denotes the $\nnode$-dimensional vector with all entries equal to $1$. 
Then we define $Y_i \in \R^{\nnode \times \nnode}$ as
\bal\nn
Y_i = \mathbf{D}_i - \mathbf{\edgewt}_i.
\eal
We perform the experiment with training samples of size $n\in\{200,500,1000\}$, 
number of nodes $\nnode = 10$, 
in different levels of noise with $a\in\{0,0.02,0.1\}$, respectively. 
\end{case}

\begin{case}\label{case:employment_composition}
(Real-world data with the compositional responses.) 
The state-level employment data from the Federal Reserve Economic Data (FRED) report employment counts in nine occupational subclasses: (1) Mining and Logging; (2) Construction; (3) Manufacturing; (4) Trade, Transportation, and Utilities; (5) Information; (6) Professional and Business Services; (7) Leisure and Hospitality; (8) Other Services; and (9) Government. 
We focus on the year 2010 and consider the $49$ states except for Hawaii, which is excluded due to missing records. 
The responses are the employment occupational compositions for each state, represented as a $9$-dimensional vector of occupational subclass shares summing to one. 
In addition, we assemble state-level economic, demographic, and geographic information in the year 2010 
from the Federal Reserve Economic Data (FRED), 
the U.S.\ Department of Labor, Wage \& Hour Division, 
and the U.S. Census Bureau, 
which turns to be $17$ predictors as listed in Table~\ref{tab:state_predictors}. 
See Section~\ref{sec:employment_data} in the \sppl for more details about the data. 
\end{case}

\begin{case}\label{case:energy_sphere}
(Real-world data with the spherical responses.) 
State-level net electricity generation data for the year 2010 for all $50$ states were obtained from the U.S.\ Energy Information Administration (EIA) Annual Electric Power Industry Report (\url{https://www.eia.gov/electricity/data.php\#generation}).
For each state, annual generation in megawatt-hours (MWh) was aggregated across twelve fuel-source categories: 
(1) Coal; (2) Hydroelectric Conventional; (3) Natural Gas; (4) Other; (5) Other Biomass; 
(6) Petroleum; (7) Wind; (8) Nuclear; (9) Other Gases; (10) Wood and Wood Derived Fuels; (11) Solar Thermal and Photovoltaic; and (12) Geothermal. 
Each state's generation profile is a $12$-dimensional compositional vector of non-negative parts summing to one. 
Following \citet{scea:14}, the componentwise square root maps each composition to the positive orthant $\sphere^{11}_+$ of the unit sphere $\sphere^{11}$, yielding sphere-valued responses $Y_i\in\sphere^{11}_+$ with $\|Y_i\|=1$. 
The predictors are the same $17$ state-level economic, demographic, and geographic variables used in Experiment~\ref{case:employment_composition}. 
\end{case}

\begin{table}[hbt!]
\centering
\caption{State-level predictors used in the employment composition analysis.}
\label{tab:state_predictors}
\begin{tabular}{rll}
\toprule
Index & Variable & Explanation \\
\midrule
1 & Unemployment rate & percentage of unemployed individuals \\
2 & Income & annual median household income \\
3 & GDP & state-level gross domestic product \\
4 & Minimum wage & state minimum hourly wage \\
5 & RPP & relative housing cost index (national average $= 100$) \\
6 & HPI & index of housing price trends based on transactions \\
7 & Business & number of new business applications \\
8 & Part-time rate & involuntary part-time employment rate \\
9 & U-6 rate  & a broader unemployment measure \\
10 & Urbanization & share of population in urban areas \\
11 & Population density & residents per square mile \\
12 & Population & total number of residents \\
13 & Sex ratio & male-to-female population ratio \\
14 & Homeownership  & share of owner-occupied housing units \\
15 & Rental & rental vacancy rate \\
16 & Education & proportion with a bachelor’s degree or higher \\
17 & Insurance coverage & uninsured population rate \\
\bottomrule
\end{tabular}
\end{table}

\subsection{Results}\label{sec:expr_results}

\begin{figure}[!hbt]
\centering 
\includegraphics[width=1.0\linewidth]{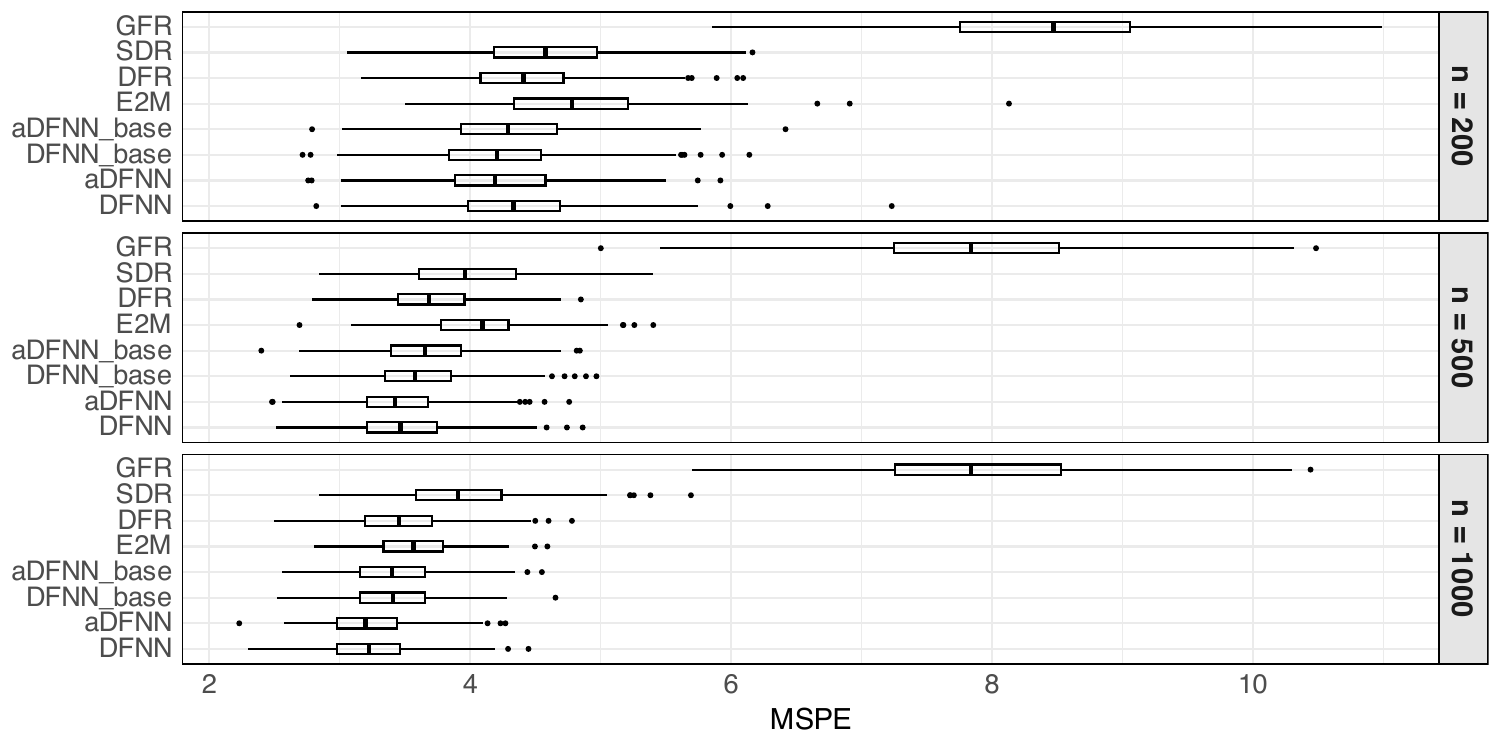}
\caption{Boxplots of MSPEs for GFR, SDR, DFR, E2M, \adfnnbase, \dfnnbase, \adfnn, and \dfnn for Experiment~\ref{case:distn_lowdim2} with sample size $n\in\{200,500,1000\}$.}
\label{fig:distn_lowdim2}
\end{figure}

\begin{figure}[!hbt]
\centering 
\includegraphics[width=1.0\linewidth]{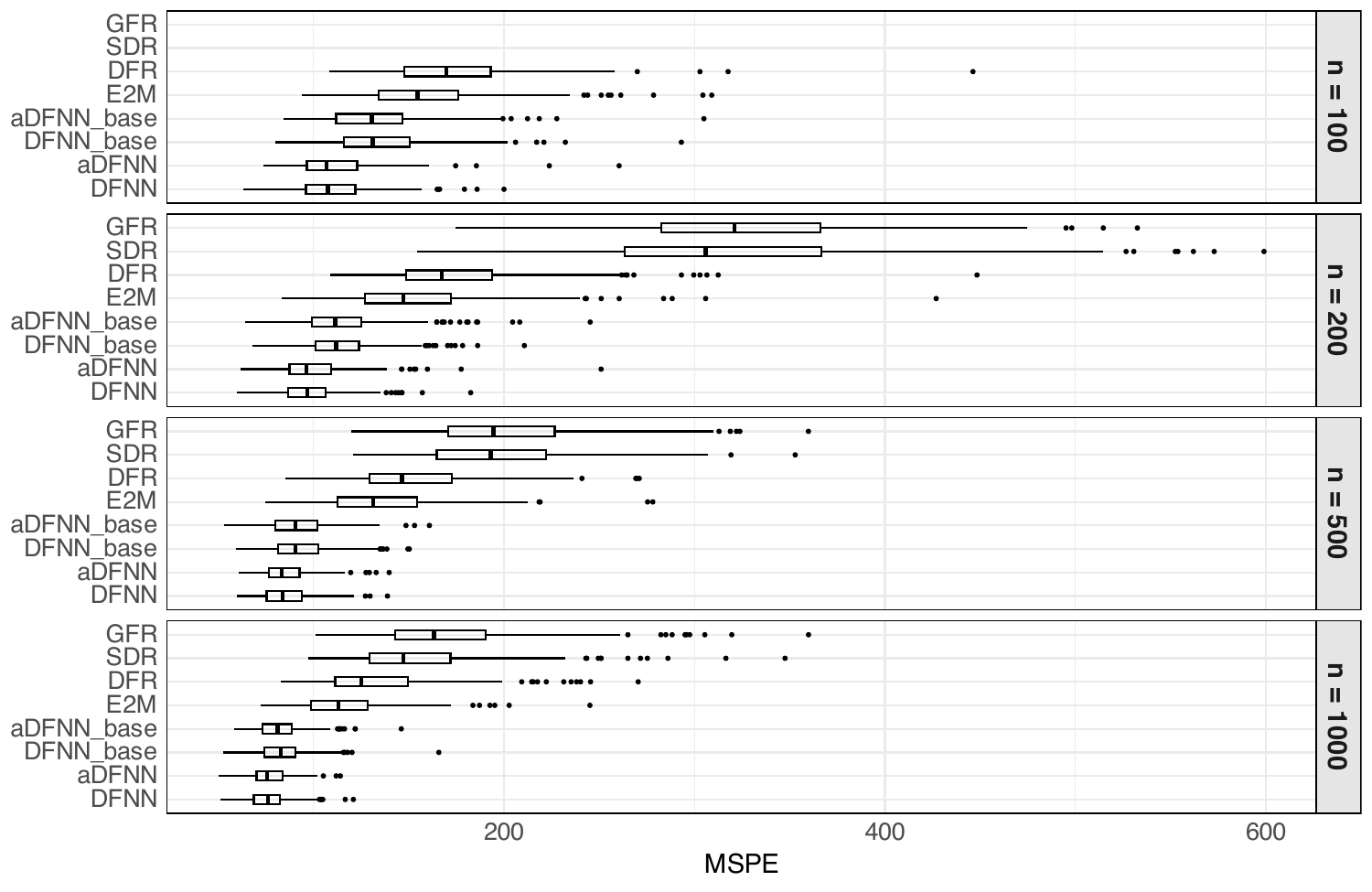}
\caption{Boxplots of MSPEs for GFR, SDR, DFR, E2M, \adfnnbase, \dfnnbase, \adfnn, and \dfnn for Experiment~\ref{case:distn_highdim} with sample size $n\in\{100,200,500,1000\}$. Outliers with MSPE values above 600 are omitted for clarity. This includes five outliers from SDR at $n=200$, one from GFR at $n=200$, and all results for GFR and SDR under the $n=100$ setting.}
\label{fig:distn_highdim}
\end{figure}

For Experiments~\ref{case:distn_lowdim2}--\ref{case:distn_highdim} with distributional responses, 
MSPEs over $\nMC=250$ Monte Carlo runs for different methods are summarized in Tables~\ref{tab:distn_lowdim2}--\ref{tab:distn_highdim} and Figures~\ref{fig:distn_lowdim2}--\ref{fig:distn_highdim}. 
Both \adfnn and \dfnn and their base variants consistently outperform all other competing methods across all considered sample sizes, demonstrating superior predictive performance in terms of both prediction accuracy and stability. 

\begin{table}[!hbt]
    \centering
    \caption{Average and standard deviation (in parentheses) of MSPEs of GFR, SDR, DFR, E2M, \adfnnbase, \dfnnbase, \adfnn, and \dfnn for Experiment~\ref{case:distn_lowdim2} with sample size $n\in\{200,500,1000\}$.}
    \label{tab:distn_lowdim2}
    \resizebox{\linewidth}{!}{
    \begin{tabular}{rllllllll}
    \toprule
    $n$ & GFR & SDR & DFR & E2M & \adfnnbase & \dfnnbase & \adfnn & \dfnn \\
    \midrule
     200  & 8.42 (0.99) & 4.63 (0.57) & 4.43 (0.51) & 4.82 (0.72) & 4.26 (0.52) & 4.23 (0.56) & \textbf{4.21} (0.50) & 4.36 (0.57) \\
     500  & 7.84 (0.93) & 3.99 (0.51) & 3.71 (0.40) & 4.07 (0.44) & 3.64 (0.41) & 3.63 (0.41) & \textbf{3.46} (0.39) & 3.49 (0.40) \\
     1000  & 7.90 (0.93) & 3.95 (0.50) & 3.47 (0.40) & 3.58 (0.36) & 3.42 (0.35) & 3.41 (0.35) & \textbf{3.22} (0.34) & 3.24 (0.35) \\
    \bottomrule
    \end{tabular}}
\end{table}

\begin{table}[!hbt]
    \centering
    \caption{Average and standard deviation (in parentheses) of MSPEs of GFR, SDR, DFR, E2M, \adfnnbase, \dfnnbase, \adfnn, and \dfnn for Experiment~\ref{case:distn_highdim} with sample size $n\in\{100,200,500,1000\}$.}
    \label{tab:distn_highdim}
    \resizebox{\linewidth}{!}{
    \begin{tabular}{rllllllll}
    \toprule
    $n$ & GFR & SDR & DFR & E2M & \adfnnbase & \dfnnbase & \adfnn & \dfnn \\
    \midrule
      100  & 38521.76 (140237.60) & 67426.65 (588414.84) & 174.70 (40.13) & 159.01 (36.08) & 133.87 (29.41) & 135.44 (28.33) & 110.80 (22.81) & \textbf{110.78} (20.71) \\
     200  & 329.75 (68.37) & 333.38 (124.27) & 175.24 (42.31) & 153.99 (40.36) & 114.88 (25.05) & 114.56 (20.64) & 99.83 (20.15) & \textbf{98.79} (16.12) \\
     500  & 201.71 (42.60) & 197.43 (41.60) & 152.90 (34.13) & 135.52 (30.76) & 92.41 (17.20) & 93.36 (16.51) & \textbf{85.47} (13.15) & 85.45 (13.81) \\
     1000  & 172.28 (40.67) & 155.25 (38.31) & 133.24 (32.10) & 116.03 (24.81) & 82.90 (12.83) & 83.67 (13.83) & \textbf{77.23} (10.74) & 77.01 (11.39) \\
    \bottomrule
    \end{tabular}}
\end{table}

\begin{figure}[!hbt]
\centering 
\includegraphics[width=1.0\linewidth]{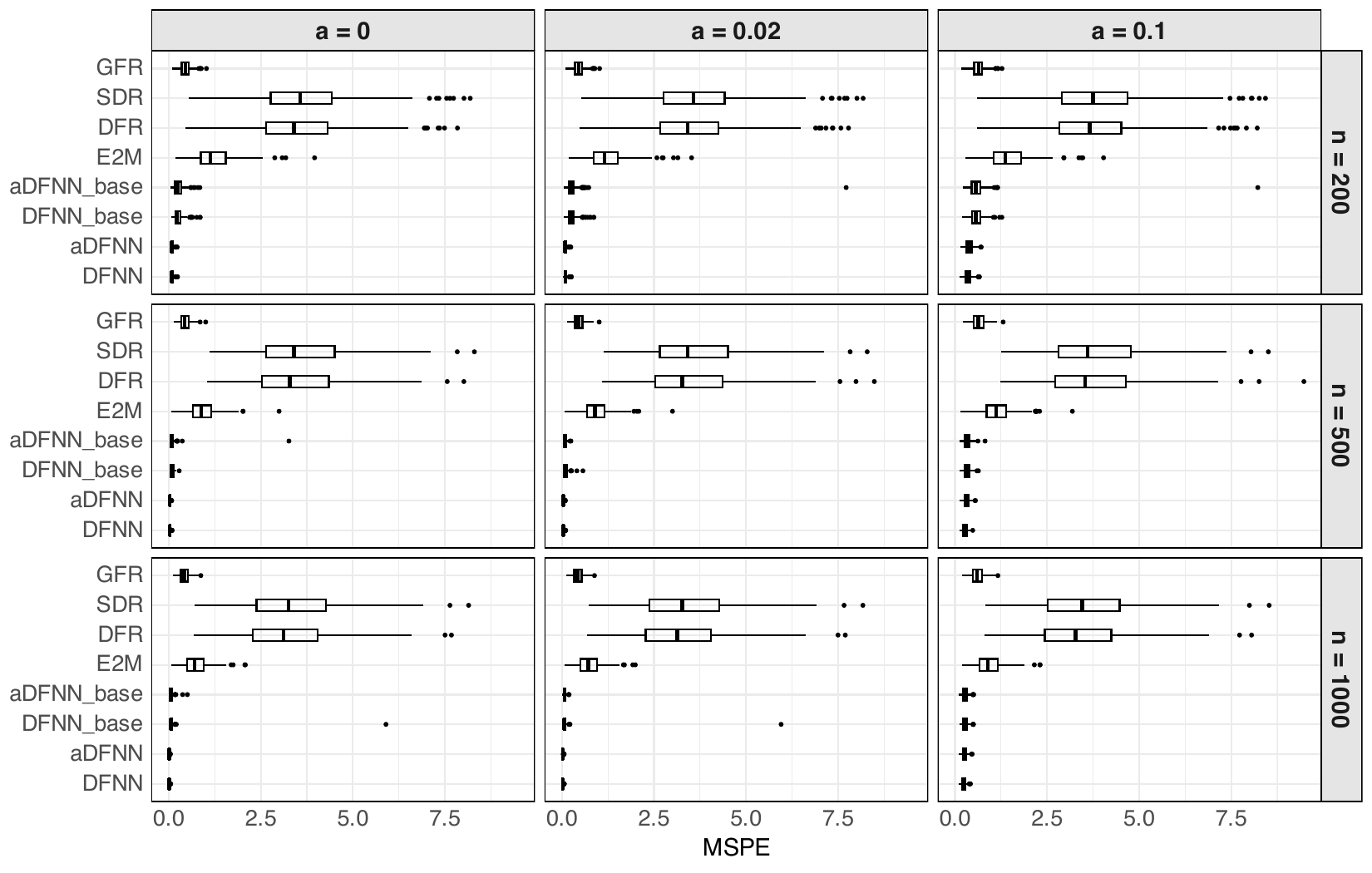}
\caption{Boxplots of MSPEs for GFR, SDR, DFR, E2M, \adfnnbase, \dfnnbase, \adfnn, and \dfnn for Experiment~\ref{case:network} with sample size $n\in\{200,500,1000\}$ and different levels of noise ($a=0$ corresponding to noiseless settings and larger $a$ corresponding to higher levels of noise). Three outliers with MSPE values above 10, all from the DFR method, occurred under the settings ($n=200, a=0.02$), ($n=1000, a=0.02$), and ($n=1000, a=0.1$), and are omitted for visualization purposes.}
\label{fig:net}
\end{figure}

For Experiment~\ref{case:network} with network-valued responses, MSPEs over $\nMC=250$ Monte Carlo runs are summarized in  Figure~\ref{fig:net} and Tables~\ref{tab:net_a0}--\ref{tab:net_a01}. 
While MSPEs increase at higher levels of noise, 
both \adfnn and \dfnn attain the lowest average errors and smallest variances across all sample sizes and for noiseless data and data contaminated with different levels of noise.

\begin{table}[!hbt]
  \centering
  \caption{Average and standard deviation (in parentheses) of MSPEs of GFR, SDR, DFR, E2M, \adfnnbase, \dfnnbase, \adfnn, and \dfnn for Experiment~\ref{case:network} in noiseless settings ($a=0$) with sample size $n\in\{200,500,1000\}$.}
  \label{tab:net_a0}
  \resizebox{\linewidth}{!}{
  \begin{tabular}{rllllllll}
    \toprule
    $n$ & GFR & SDR & DFR & E2M & \adfnnbase & \dfnnbase & \adfnn & \dfnn \\
    \midrule
    200  & 0.459 (0.151) & 3.725 (1.419) & 3.574 (1.367) & 1.228 (0.554) & 0.266 (0.133) & 0.254 (0.124) & \textbf{0.085} (0.033) & 0.091 (0.036) \\
    500  & 0.454 (0.144) & 3.637 (1.308) & 3.490 (1.274) & 0.924 (0.402) & 0.103 (0.206) & 0.090 (0.046) & \textbf{0.028} (0.013) & 0.033 (0.015) \\
    1000 & 0.427 (0.143) & 3.352 (1.311) & 3.193 (1.261) & 0.743 (0.332) & 0.067 (0.049) & 0.087 (0.371) & \textbf{0.011} (0.007) & 0.016 (0.008) \\
    \bottomrule
  \end{tabular}}
\end{table}

\begin{table}[!hbt]
  \centering
  \caption{Average and standard deviation (in parentheses) of MSPEs of GFR, SDR, DFR, E2M, \adfnnbase, \dfnnbase, \adfnn, and \dfnn for Experiment~\ref{case:network} in settings of low level of noise ($a=0.02$) with sample size $n\in\{200,500,1000\}$.}
  \label{tab:net_a002}
  \resizebox{\linewidth}{!}{
  \begin{tabular}{rllllllll}
    \toprule
    $n$ & GFR & SDR & DFR & E2M & \adfnnbase & \dfnnbase & \adfnn & \dfnn \\
    \midrule
    200  & 0.466 (0.152) & 3.734 (1.422) & 3.617 (1.487) & 1.232 (0.542) & 0.305 (0.488) & 0.270 (0.130) & \textbf{0.096} (0.036) & 0.103 (0.039) \\
    500  & 0.461 (0.146) & 3.645 (1.309) & 3.502 (1.307) & 0.938 (0.403) & 0.098 (0.045) & 0.102 (0.059) & \textbf{0.040} (0.016) & 0.043 (0.017) \\
    1000 & 0.435 (0.145) & 3.360 (1.315) & 3.526 (5.236) & 0.750 (0.323) & 0.072 (0.037) & 0.096 (0.374) & \textbf{0.023} (0.009) & 0.025 (0.010) \\
    \bottomrule
  \end{tabular}}
\end{table}

\begin{table}[!hbt]
  \centering
  \caption{Average and standard deviation (in parentheses) of MSPEs of GFR, SDR, DFR, E2M, \adfnnbase, \dfnnbase, \adfnn, and \dfnn for Experiment~\ref{case:network} in settings of high level of noise ($a=0.1$) with sample size $n\in\{200,500,1000\}$.}
  \label{tab:net_a01}
  \resizebox{\linewidth}{!}{
  \begin{tabular}{rllllllll}
    \toprule
    $n$ & GFR & SDR & DFR & E2M & \adfnnbase & \dfnnbase & \adfnn & \dfnn \\
    \midrule
    200  & 0.649 (0.191) & 3.928 (1.466) & 3.798 (1.415) & 1.449 (0.596) & 0.614 (0.515) & 0.581 (0.174) & 0.384 (0.104) & \textbf{0.357} (0.098) \\
    500  & 0.643 (0.186) & 3.831 (1.345) & 3.733 (1.369) & 1.135 (0.442) & 0.327 (0.093) & 0.327 (0.093) & 0.312 (0.084) & \textbf{0.273} (0.071) \\
    1000 & 0.608 (0.185) & 3.537 (1.357) & 3.538 (2.714) & 0.929 (0.359) & 0.264 (0.077) & 0.265 (0.077) & 0.261 (0.068) & \textbf{0.231} (0.059) \\
    \bottomrule
  \end{tabular}}
\end{table}

\begin{figure}[!hbt]
\centering 
\includegraphics[width=1.0\linewidth]{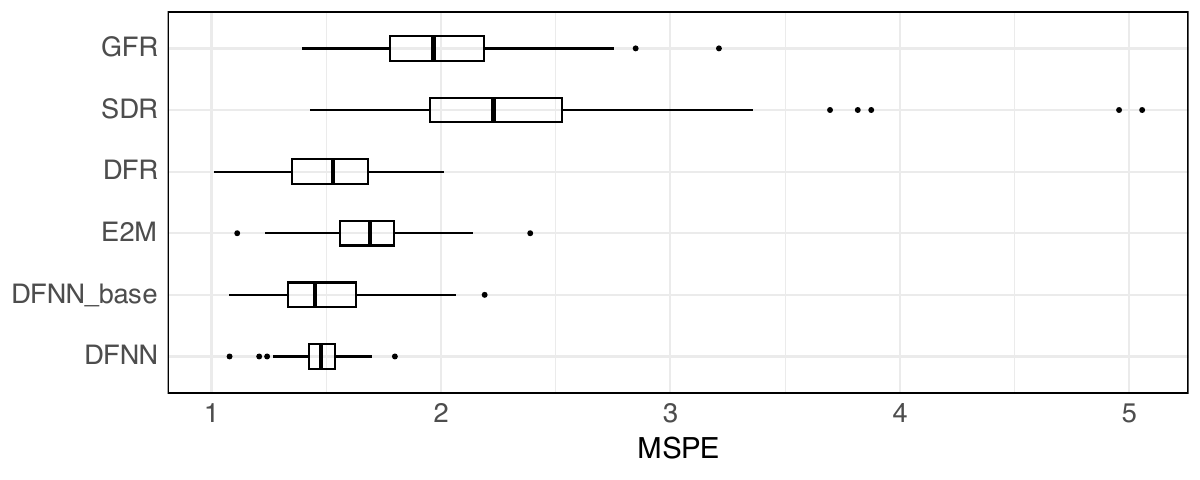}
\caption{Boxplots of MSPEs from $100$ times $10$-fold cross-validation for GFR, SDR, DFR, E2M, \dfnnbase, and \dfnn for Experiment~\ref{case:employment_composition}.}
\label{fig:employment_composition}
\end{figure}

\begin{figure}[!hbt]
\centering
\includegraphics[width=0.8\linewidth]{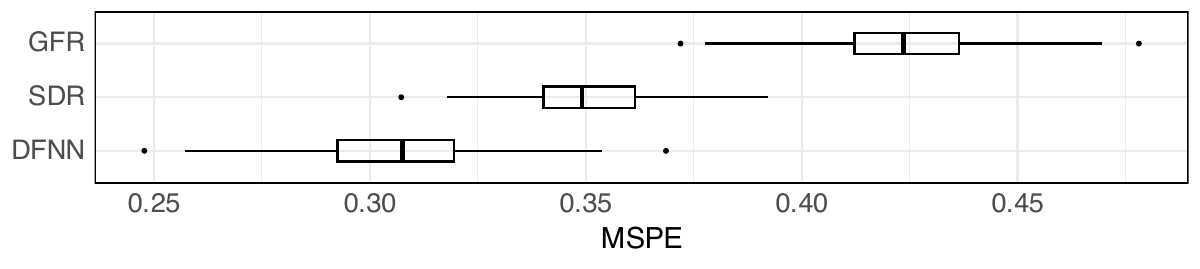}
\caption{Boxplots of MSPEs from $100$ times $10$-fold cross-validation for GFR, SDR, and \dfnn for Experiment~\ref{case:energy_sphere}.}
\label{fig:sphere_energy}
\end{figure}

For the real-data applications (Experiments~\ref{case:employment_composition} and~\ref{case:energy_sphere}), MSPEs from $100$ Monte Carlo runs 
of $10$-fold cross-validation are summarized in Figures~\ref{fig:employment_composition}--\ref{fig:sphere_energy} and Tables~\ref{tab:employment_composition}--\ref{tab:sphere_energy}.
In both experiments, \dfnn outperforms all competing methods. 

\begin{table}[!hbt]
  \centering
  \caption{Average and standard deviation (in parentheses) of MSPEs of GFR, SDR, DFR, E2M, \dfnnbase, and \dfnn for Experiment~\ref{case:employment_composition}.}
  \label{tab:employment_composition}
  \begin{tabular}{llllll}
    \toprule
     GFR & SDR & DFR & E2M & \dfnnbase & \dfnn \\
    \midrule
     2.02 (0.32) & 2.34 (0.61) & 1.53 (0.22) & 1.68 (0.20) & 1.49 (0.23) & \textbf{1.48} (0.11) \\
    \bottomrule
  \end{tabular}
\end{table}

\begin{table}[!hbt]
  \centering
  \caption{Average and standard deviation (in parentheses) of MSPEs of GFR, SDR, and \dfnn for Experiment~\ref{case:energy_sphere}.}
  \label{tab:sphere_energy}
  \begin{tabular}{lll}
    \toprule
    GFR & SDR & \dfnn \\
    \midrule
    0.42 (0.02) & 0.35 (0.02) & \textbf{0.31} (0.02) \\
    \bottomrule
  \end{tabular}
\end{table}

\section{Discussion}\label{sec:disc}

This work proposes DFNNs as an end-to-end framework for regression with responses in metric spaces, combining a weighted \F mean output layer with standard hidden-layer representations and training via empirical \F risk minimization. For the special case of Euclidean responses, the proposed DFNNs reduce to classical DNNs. We establish a universal approximation result showing that, under mild regularity conditions, DFNNs can approximate the conditional \F mean up to arbitrary accuracy. Beyond approximation, we develop a statistical theory for DFNNs by establishing uniform consistency and convergence rates for the empirical DFNN output layer over the predictor domain and neural network parameter space, and by deriving generalization guarantees with explicit bounds on excess \F risk and prediction error, including rates under both norm-based and parameter-count-based complexity controls. These results provide a theoretical foundation for deep learning with metric-space-valued responses that complements the expressive power established by the universal approximation theorem. We also show through extensive empirical experiments that DFNNs consistently outperform state-of-the-art methods across distributional, network-valued, and compositional response settings.   Future work includes principled metric selection, extending the theory to broader architectural regimes, and analyzing the algorithmic aspects of DFNN training.

\references

\appendix

\section{Detailed Review of Related Work}
\label{sec:related}

\subsection{\F Mean} The notion of \F means or barycenters is a generalization of averages to random objects residing in a general metric space. 
It is defined as the minimizer(s) of a \F functional---the expected squared distance to the random object of interest---over the corresponding metric space \citep{frec:48}. 
The existence and uniqueness of \F means is guaranteed for Hadamard spaces \citep{stur:03} and have been studied for Riemannian manifolds, Wasserstein space, Alexandrov spaces, and proper metric spaces \citep[e.g.,][]{ague:11,ohta:12,kim:17,lego:17}. 
Properties including asymptotic behavior of empirical barycenters have been established for Riemannian manifolds, Wasserstein space and general metric spaces \citep[e.g.,][]{bhat:02,scho:19,zeme:19,ahid:20,lego:22,brun:24}. 
As an extension to a supervised setting where the metric-space-valued random object is observed along with Euclidean predictors, the question of interest is to estimate the \emph{conditional \F mean} of the random object given the predictors. 
Specifically, the conditional \F mean is defined as the minimizer(s) of the expected squared distance to the random object conditioning on the predictors, 
which is the target of regression with metric-space-valued responses and Euclidean predictors \citep{pete:19}.

\subsection{\F Regression for Metric-Space-Valued Responses}
First introduced by \cite{pete:19}, \F regression is a class of methods that aims to estimate the conditional \F mean of a response lying in a metric space given Euclidean predictors. 
Global \F regression and local \F regression by can be viewed as generalizations of multiple linear models and local linear regression and uniform rates of convergence have been established for the corresponding estimators of the conditional \F means \citep{pete:19,chen:22}. 
Since then, other variants of \F regression have been developed as well as methods to improve the prediction performance of global \F regression \citep[][among others]{scho:20,lin:21:5,bhat:23:3,ghos:23:2,song:23:2,capi:24,qiu:24,yan:24,bhat:25,kuri:25:3,zhou:25:3}. 

Beside methodologies---such as \F regression---that are developed for responses lying in a general metric space, there are also regression approaches that are specialized for a specific type of non-Euclidean responses, i.e., responses lying in a specific metric space, paired with Euclidean predictors. 
While they may be only applicable to one type of non-Euclidean responses, thereby lacking generality compared with methodologies developed for responses lying in a general metric space, specialized methods exploit structure of particular response spaces and hence may achieve better predictive accuracy and provide more interpretable results than approaches applicable to general metric spaces. 

\subsection{Regression Methods for Finite-Dimensional Riemannian Responses}
The analysis of random objects taking values in a finite-dimensional differentiable Riemannian manifold, including statistics for manifold-valued data---which foster object-oriented data analysis---shape analysis, and geometric statistics, is a topic that has been extensively studied \citep[e.g.,][]{marr:14,dryd:16,marr:21,huck:21}. 
Regression models for this type of non-Euclidean responses paired with Euclidean predictors include regression for general Riemannian responses \citep[e.g.,][]{davi:07,shi:09,hink:12,corn:17,lin:17:2,choi:25}, 
spherical/directional regression \citep[e.g.,][]{fish:87,jupp:87,chan:89,pren:89,fish:95,mard:09,di:14,scea:19}, 
regression for SPD matrices \citep[e.g.,][]{zhu:09,yuan:12,pete:19:3,lin:23:3,xu:25}, 
and network/graph regression \citep[e.g.,][]{seve:21,cali:22,seve:22,zhou:22:4,zhou:23}, among others. 

\subsection{Regression Methods for Compositional Responses} 
Compositional data are another type of data which cannot be treated as classical Euclidean data due to the constraints that entries have to be non-negative and sum up to $1$. 
Methods for the analysis of compositional data including regression with compositional responses have been developed mainly using Aitchison metrics including the simple pairwise log-ratio (LR),  additive log-ratio (ALR), and  centered log-ratio (CLR) metrics \citep[e.g.,][]{aitc:82,aitc:86,filz:18,gree:21,gree:23} and geodesic metric on a (hyper)sphere applied to square root transformed compositions \citep[e.g.,][]{scea:11,scea:17,li:23}. 

\subsection{Regression Methods for Distributional Responses} 
Probability distributions or probability density functions are another important type of non-Euclidean data \citep{pete:22}. 
While the space of distributions is equipped with Riemannian structures under certain metrics, methods for finite-dimensional Riemannian data cannot be directly applied to the analysis of distributional data due to their infinite-dimensional nature.  
For regression with univariate distributional responses, existing methods include: 
plain vanilla methods that view density functions as elements in an $\ltwo$ space---a space of square-integrable functions on certain domain \citep[e.g.,][]{oliv:13}; 
transformation- or embedding-based methods, which first map distributions into an unconstrained Hilbert space using transformations including log quantile density (LQD) centered log-ratio (CLR) or kernel mean embedding so that classical functional regression methods can then be applied \citep[][among others]{vand:14,pete:16:1,hron:16,szab:16,pete:19:8,han:20,maie:25}; 
manifold-based methods, which leverages the Riemannian or pseudo-Riemannian geometries of the space of distributions endowed with the Fisher--Rao metric or $2$-Wasserstein metric \citep[][among others]{sriv:07,bigo:17:2,caze:18,ghod:22:2,pego:22,zhan:21:3,chen:19:4,zhu:23:3,zhou:24:2,zhu:24}. 
In addition to univariate distributional responses, regression methods have also been developed for multivariate distributional responses
\citep[][among others]{gueg:18,chen:23,ghod:23:2,hron:23,fan:24,okan:24,zhu:24}.

\subsection{Dimension Reduction for \F Regression} 
To address the challenges posed by high-dimensional predictors in learning metric-space-valued responses, a number of works have extended classical dimension reduction techniques to the \F regression setting. These include sufficient dimension reduction \citep{dong:22,virt:22,ying:22,zhan:24:2,weng:25,ying:25}, single index models \citep{ghos:23:2,bhat:23:3}, and principal component regression \citep{song:23:2}, among others. 

\subsection{Deep Learning Based Methods for Regression for Non-Euclidean Data} 
Deep learning based methods developed for Euclidean data have been generalized to non-Euclidean data. 
Arising from representation learning for symbolic data, 
manifold considerations are found to be necessary to capture the structural geometry of some complex data \citep{nick:17}. 
Existing work of extending deep learning based methods to regression or supervised learning with non-Euclidean data mostly tackle non-Euclidean inputs. Among that, most investigations focus on a specific type of Riemannian manifolds, such as data in hyperbolic spaces \citep[e.g.,][]{gane:18,gulc:19,liu:19,cham:19,chen:20:7,peng:21}, SPD matrices \citep[e.g.,][]{huan:17,broo:19,nguy:19:2,nguy:21}, spherical data \citep[e.g.,][]{cohe:18,coor:18,fang:20:2}, 
while representation learning has also been investigated for non-Euclidean data, in particular Riemannian data, which is often referred to as geometric (deep) learning \citep[e.g.,][]{masc:15,bron:17,kipf:17,chak:20,sun:24}. 
Another line of research deals with regression with manifold-valued outputs \citep[e.g.,][]{pan:22,chen:23:3}. 
A more recent line of work integrates representation learning into the \F regression framework for responses in general metric spaces paired with Euclidean predictors. \cite{iao:25} proposed a deep learning based method that assumes the regression function---namely, the conditional \F means---resides on a low-dimensional manifold. 
Their approach employs a DNN to learn this manifold representations, on which local \F regression is then applied to predict responses in the metric space. 
Nonetheless, its effectiveness relies critically on the low-dimensional manifold representation assumption on the conditional \F mean, and, inherent from local \F regression, it remains vulnerable to the curse of dimensionality. 

\subsection{Related Learning Frameworks for Different Purposes}
There exist other learning frameworks that are related to the analysis of non-Euclidean data, but they---at least alone---are not methods for regression with non-Euclidean responses paired with Euclidean predictors. Examples include: 
manifold learning \citep{ghoj:23}, which can be used to embed Euclidean inputs into a low-dimensional manifold; 
metric learning \citep{bell:15}, which learns similarity/distance for a specific supervised or unsupervised task, e.g., for classification or dimension reduction. 

\section{Choices of Hyperparameters for DFNN}\label{sec:hyper}

\subsection{Choices of Architectural and Regularization-related Hyperparameters for DFNN} \label{sec:arch_rgl}

For the proposed DFNN method, the grids of candidate values for each architectural and regularization-related hyperparameters for each experiment are listed in Tables~\ref{tab:grid-hp_exp1}--\ref{tab:grid-hp_exp5},
and the selected values via grid search for each experiment are listed in Tables~\ref{tab:adfnn-hp}--\ref{tab:base-hp}.

\begin{table}[!hbt]
\centering
\caption{Grids of architectural and regularization-related hyperparameters in Experiment~\ref{case:distn_lowdim2}.}
\label{tab:grid-hp_exp1}
\begin{tabular}{ll}
\toprule
Hyperparameter & Values \\
\midrule
Hidden sizes $\{\wlyr_l\}_{l=1}^{\nlyr-1}$  &  512, 1024, 2048 \\
Hidden size $\wlyr_{\nlyr}$   & 8, 16, 24 \\
Hidden width $\nlyr$ & 3, 4 \\
LR & 0.001, 0.005, 0.01 \\
DR  & 0.1, 0.2, 0.3 \\
\bottomrule
\end{tabular}
\end{table}

\begin{table}[!hbt]
\centering
\caption{Grids of architectural and regularization-related hyperparameters in Experiment~\ref{case:distn_highdim}.}
\label{tab:grid-hp_exp2}
\begin{tabular}{ll}
\toprule
Hyperparameter & Values \\
\midrule
Hidden sizes $\{\wlyr_l\}_{l=1}^{\nlyr-1}$  &  1024, 2048, 4096 \\
Hidden size $\wlyr_{\nlyr}$   &  16, 24, 32 \\
Hidden width $\nlyr$ & 3, 4 \\
LR & 0.001, 0.005, 0.01 \\
DR  & 0.1, 0.2, 0.3 \\
\bottomrule
\end{tabular}
\end{table}

\begin{table}[!hbt]
\centering
\caption{Grids of architectural and regularization-related hyperparameters in Experiment~\ref{case:network}.}
\label{tab:grid-hp_exp3}
\begin{tabular}{ll}
\toprule
Hyperparameter & Values \\
\midrule
Hidden sizes $\{\wlyr_l\}_{l=1}^{\nlyr-1}$   & 256, 512, 1024, 4096 \\
Hidden size $\wlyr_{\nlyr}$   & 5, 10, 15  \\
Hidden width $\nlyr$ & 3, 4 \\
LR & 0.001, 0.01, 0.1 \\
DR  & 0.1, 0.2, 0.3 \\
\bottomrule
\end{tabular}
\end{table}

\begin{table}[!hbt]
\centering
\caption{Grids of architectural and regularization-related hyperparameters in Experiment~\ref{case:employment_composition}.}
\label{tab:grid-hp_exp4}
\begin{tabular}{ll}
\toprule
Hyperparameter & Values \\
\midrule
Hidden sizes $\{\wlyr_l\}_{l=1}^{\nlyr-1}$  & 128, 256, 512, 1024 \\
Hidden size $\wlyr_{\nlyr}$  & 7, 14, 21 \\
Hidden width $\nlyr$ & 3, 4 \\
LR & 0.001, 0.01, 0.01 \\
DR  & 0.1, 0.2, 0.3 \\
\bottomrule
\end{tabular}
\end{table}

\begin{table}[!hbt]
\centering
\caption{Grids of architectural and regularization-related hyperparameters in Experiment~\ref{case:energy_sphere}.}
\label{tab:grid-hp_exp5}
\begin{tabular}{ll}
\toprule
Hyperparameter & Values \\
\midrule
Hidden sizes $\{\wlyr_l\}_{l=1}^{\nlyr-1}$  & 256, 512, 1024 \\
Hidden size $\wlyr_{\nlyr}$  & 8, 16, 24 \\
Hidden width $\nlyr$ & 3, 4 \\
LR & 0.001, 0.01, 0.1 \\
DR  & 0.1, 0.2, 0.3 \\
\bottomrule
\end{tabular}
\end{table}

\begin{table}[!hbt]
\centering
\caption{Architectural and regularization-related hyperparameters for \adfnn.}
\label{tab:adfnn-hp}
\begin{tabular}{lccccc}
\toprule
Experiment & Hidden sizes $\{\wlyr_l\}_{l=1}^{\nlyr-1}$ & Hidden width $\nlyr$ & Hidden size $\wlyr_{\nlyr}$ & LR & DR \\
\midrule
Experiment~\ref{case:distn_lowdim2} & $2048$ & $4$ & $8$ & $0.001$ & $0.3$ \\
Experiment~\ref{case:distn_highdim} & $4096$ & $4$ & $32$ & $0.005$ & $0.3$ \\
Experiment~\ref{case:network} & $4096$ & $4$ & $15$ & $0.01$ & $0.1$ \\
\bottomrule
\end{tabular}
\end{table}

\begin{table}[!hbt]
\centering
\caption{Architectural and regularization-related hyperparameters for \dfnn.}
\label{tab:dfnn-hp}
\begin{tabular}{lccccc}
\toprule
Experiment & Hidden sizes $\{\wlyr_l\}_{l=1}^{\nlyr-1}$ & Hidden width $\nlyr$ & Hidden size $\wlyr_{\nlyr}$ & LR & DR \\
\midrule
Experiment~\ref{case:distn_lowdim2} & $2048$ & $3$ & $8$ & $0.001$ & $0.3$ \\
Experiment~\ref{case:distn_highdim} & $4096$ & $3$ & $32$ & $0.005$ & $0.3$ \\
Experiment~\ref{case:network} & $4096$ & $3$ & $15$ & $0.01$ & $0.3$ \\
Experiment~\ref{case:employment_composition} & $1024$ & $4$ & $14$ & $0.01$ & $0.3$ \\
Experiment~\ref{case:energy_sphere} & $1024$ & $3$ & $24$ & $0.01$ & $0.2$ \\
\bottomrule
\end{tabular}
\end{table}

\begin{table}[!hbt]
\centering
\caption{Architectural and regularization-related hyperparameters for \adfnnbase\ and \dfnnbase.}
\label{tab:base-hp}
\begin{tabular}{lccccc}
\toprule
Experiment & Hidden sizes $\{\wlyr_l\}_{l=1}^{\nlyr-1}$ & Hidden width $\nlyr$ & Hidden size $\wlyr_{\nlyr}$ & LR & DR \\
\midrule
Experiment~\ref{case:distn_lowdim2} & $64$ & $4$ & $4$ & $0.005$ & $0.3$ \\
Experiment~\ref{case:distn_highdim} & $64$ & $4$ & $16$ & $0.005$ & $0.3$ \\
Experiment~\ref{case:network} & $32$ & $4$ & $10$ & $0.005$ & $0.0$ \\
Experiment~\ref{case:employment_composition} & $32$ & $4$ & $7$ & $0.005$ & $0.3$ \\
\bottomrule
\end{tabular}
\end{table}

\subsection{Choices of Hyperparameters Used in the Early Stopping Rule} \label{sec:earlystop}

For each case in the experiments, we select the values of the hyperparameters, $\nepint$, $\tolr$, and $\npatn$, used in the early stopping rule stated in Section~\ref{sec:impl_details}, via a grid search and then use the corresponding choices for the experiment. 
The values selected for each experiment are listed in Tables~\ref{tab:earlystop_adfnn}--\ref{tab:earlystop_base}. 

\begin{table}[!hbt]
\centering
\caption{Early-stopping hyperparameters used in the experiments for \adfnn.}
\label{tab:earlystop_adfnn}
\begin{tabular}{lccc}
\toprule
Experiment & $\nepint$ & $\tolr$ & $\npatn$ \\
\midrule
Experiment~\ref{case:distn_lowdim2} & $50$ & $1.0\times10^{-5}$ & $150$ \\
Experiment~\ref{case:distn_highdim} & $50$ & $1.0\times10^{-5}$ & $100$ \\
Experiment~\ref{case:network}                & $50$ & $1.0\times10^{-5}$ & $100$ \\
\bottomrule
\end{tabular}
\end{table}

\begin{table}[!hbt]
\centering
\caption{Early-stopping hyperparameters used in the experiments for \dfnn.}
\label{tab:earlystop_dfnn}
\begin{tabular}{lccc}
\toprule
Experiment & $\nepint$ & $\tolr$ & $\npatn$ \\
\midrule
Experiment~\ref{case:distn_lowdim2} & $50$ & $1.0\times10^{-5}$ & $150$ \\
Experiment~\ref{case:distn_highdim} & $50$ & $1.0\times10^{-5}$ & $100$ \\
Experiment~\ref{case:network} & $200$ & $1.0\times10^{-5}$ & $100$ \\
Experiment~\ref{case:employment_composition} & $500$ & $1.0\times10^{-5}$ & $1000$ \\
Experiment~\ref{case:energy_sphere} & $500$ & $1.0\times10^{-5}$ & $100$ \\
\bottomrule
\end{tabular}
\end{table}

\begin{table}[!hbt]
\centering
\caption{Early-stopping hyperparameters used in the experiments for \adfnnbase\ and \dfnnbase.}
\label{tab:earlystop_base}
\begin{tabular}{lccc}
\toprule
Experiment & $\nepint$ & $\tolr$ & $\npatn$ \\
\midrule
Experiment~\ref{case:distn_lowdim2} & $150$ & $1.0\times10^{-5}$ & $300$ \\
Experiment~\ref{case:distn_highdim} & $10$ & $1.0\times10^{-5}$ & $50$ \\
Experiment~\ref{case:network}                & $50$ & $1.0\times10^{-5}$ & $100$ \\
Experiment~\ref{case:employment_composition} & $50$ & $1.0\times10^{-5}$ & $100$ \\
\bottomrule
\end{tabular}
\end{table}

\section{Other Details of the Experiments}
\label{sec:expr_others}

\subsection{Other Details of Implementations of Competing Methods} \label{sec:impl_others}

For SDR and DFR, we follow the implementations in \citet{iao:25}. 
Specifically, local Fr\'echet regression is applied with a Gaussian kernel and bandwidth given by \(10\%\) of the empirical range of each input variable. 
The hyperparameters of DFR used in each setting are summarized in Table~\ref{tab:DFR-hp}. 

\begin{table}[!hbt]
\centering
\caption{Architectural and regularization-related hyperparameters for DFR.}
\label{tab:DFR-hp}
\begin{tabular}{lccccc}
\toprule
Experiment & Hidden sizes $\{\wlyr_l\}_{l=1}^{\nlyr-1}$ & Hidden width $\nlyr$ & Hidden size $\wlyr_{\nlyr}$ & LR & DR \\
\midrule
Experiment~\ref{case:distn_lowdim2} & $64$ & $4$ & $2$ & $0.0005$ & $0.3$ \\
Experiment~\ref{case:distn_highdim} & $64$ & $4$ & $2$ & $0.0005$ & $0.3$ \\
Experiment~\ref{case:network} & $32$ & $4$ & $2$ & $0.0005$ & $0.3$ \\
Experiment~\ref{case:employment_composition} & $32$ & $4$ & $2$ & $0.0005$ & $0.3$ \\
\bottomrule
\end{tabular}
\end{table}

For Experiment~\ref{case:distn_highdim} with sample size $n=100$, the original implementation of GFR, SDR, and DFR does not work as there are $\dimX=100$ predictors, i.e., $\dimX=n$, and the sample covariance matrices of the predictors are often degenerate. 
We have to modify the implementation of GFR, SDR, and DFR to make them applicable. 
In particular, all these three methods, we replace the inverse of sample covariance matrix of the predictors by the Moore--Penrose inverse. 
Moreover, for local \F regression used for SDR, we use the bandwidth chosen by cross validation. 

\subsection{U.S.\ States' Employment Occupational Compositions Used in Experiment~\ref{case:employment_composition}} \label{sec:employment_data}

State-level employment counts by occupational subclass for the year 2010 were obtained through the Federal Reserve Economic Data (FRED) platform (\url{https://fred.stlouisfed.org/}), which hosts the Bureau of Labor Statistics Quarterly Census of Employment and Wages data. For each of the $49$ continental states (Hawaii excluded due to missing records), employment counts across nine occupational subclasses---Mining and Logging; Construction; Manufacturing; Trade, Transportation, and Utilities; Information; Professional and Business Services; Leisure and Hospitality; Other Services; and Government---were converted to proportions summing to one, yielding the $9$-dimensional compositional responses.

The $17$ state-level predictors listed in Table~\ref{tab:state_predictors} were assembled from multiple publicly available U.S. government sources. All values correspond to the year 2010. Table~\ref{tab:predictor_sources} provides detailed source information for each predictor. All $17$ predictors were min-max normalized to $[0,1]$ prior to model fitting.

\begin{table}[hbt!]
\centering
\caption{Data sources for the predictors in Table~\ref{tab:state_predictors} of Section~\ref{sec:expr_setups}, 
where Census = U.S. Census Bureau; DOL = U.S. Department of Labor, Wage \& Hour Division; FRED = Federal Reserve Economic Data. 
URLs are listed for Alabama as example; URLs for other states can be obtained by replacing the state identifiers.}
\label{tab:predictor_sources}
\begin{tabular}{rllp{7cm}}
\toprule
Index & Predictor & Source & URL \\
\midrule
1  & Unemployment rate     & FRED   & \url{fred.stlouisfed.org/series/LAUST010000000000003A} \\
2  & Income                & FRED   & \url{fred.stlouisfed.org/series/MEHOINUSALA646N} \\
3  & GDP                   & FRED   & \url{fred.stlouisfed.org/series/ALNGSP} \\
4  & Minimum wage          & DOL    & \url{dol.gov/agencies/whd/state/minimum-wage/history} \\
5  & RPP                   & FRED   & \url{fred.stlouisfed.org/series/ALRPPSERVERENT} \\
6  & HPI                   & FRED   & \url{fred.stlouisfed.org/series/ALSTHPI} \\
7  & Business applications & FRED   & \url{fred.stlouisfed.org/series/BUSAPPWNSAAL} \\
8  & Part-time rate        & FRED   & \url{fred.stlouisfed.org/series/INVOLPTEMPAL} \\
9  & U-6 rate              & FRED   & \url{fred.stlouisfed.org/series/U6UNEM6AL} \\
10 & Urbanization          & Census & \url{www2.census.gov/geo/docs/reference/ua/PctUrbanRural_State.xls} \\
11 & Population density    & Census & (same as row 10) \\
12 & Population            & FRED   & \url{fred.stlouisfed.org/series/ALPOP} \\
13 & Sex ratio             & Census & \url{data.census.gov} \\
14 & Homeownership         & FRED   & \url{fred.stlouisfed.org/series/ALHOWN} \\
15 & Rental                & FRED   & \url{fred.stlouisfed.org/series/ALRVAC} \\
16 & Education             & FRED   & \url{fred.stlouisfed.org/series/GCT1502AL} \\
17 & Insurance coverage    & FRED   & \url{fred.stlouisfed.org/series/ALHICCOVPCT} \\
\bottomrule
\end{tabular}
\end{table}

\section{Gradient Computation of the Empirical Fr\'echet Risk}\label{sec:gradient_comp}

Training a DFNN via stochastic gradient descent (SGD) requires computing the gradient of the empirical Fr\'echet risk $\efrisk\{\predFNNhat(\cdot;\wtset_{\nlyr})\} 
=
\frac{1}{n} \sum_{i=1}^n d^2\{Y_i,\predFNNhat(\vX_i;\wtset_{\nlyr})\}$ in Eq.~\eqref{eq:dfnn_risk_emp} for any given $\wtset_\nlyr$, 
where $\predFNNhat(\vX_i;\wtset_{\nlyr})$ is the minimizer of the weighted Fr\'echet objective function $\ofFNNhat(\omega,\vX_i;\wtset_{\nlyr})$ over $\omega\in\msp$, as per Eq.~\eqref{eq:dfnn_output_layer_emp}. 
For $i,j\in\{1,\dots,n\}$, let 
\bal\label{eq:gfrwtij}
\widehat{w}_{ij} = \gfrwthat\{\funlyr_\nlyr(\vX_j;\wtset_\nlyr),\funlyr_\nlyr(\vX_i;\wtset_\nlyr)\}
\eal 
denote the weights as defined in Eq.~\eqref{eq:dfnn_output_weight_emp}. 
Since ReLU activation is used, the gradient of $\widehat{w}_{ij}$ with respect to $\wtset_\nlyr$ can be handled by standard automatic differentiation. 
However, differentiating through the minimization in Eq.~\eqref{eq:dfnn_output_layer_emp} needs to be addressed for each metric space $(\msp,d)$ specifically depending on its geometry.
We address each of the four metric spaces appearing in the experiments in Section~\ref{sec:empircal}, respectively. 
Specifically, the $2$-Wasserstein space of univariate distributions and the space of graph Laplacian matrices endowed with the Frobenius metric, in Experiments~\ref{case:distn_lowdim2}--\ref{case:distn_highdim} and \ref{case:network}, 
are isometric to closed convex subsets of a Hilbert space, and their weighted Fr\'echet means are obtained by a metric projection that is only piecewise smooth (Sections~\ref{sec:gradient_wasserstein}--\ref{sec:gradient_laplacian} below). 
The Aitchison simplex in Experiment~\ref{case:employment_composition} is isometric to a linear subspace of $\real^9$ via the CLR transformation, so no projection is needed and $\predFNNhat(\vX_i;\wtset_{\nlyr})$ has closed form solution (Section~\ref{sec:gradient_compositional} below). 
The sphere in Experiment~\ref{case:energy_sphere} is a Riemannian manifold, on which the weighted Fr\'echet mean $\predFNNhat(\vX_i;\wtset_{\nlyr})$ is computed by Karcher flow, 
with gradients obtained from the implicit function theorem (Section~\ref{sec:gradient_sphere} below).

\subsection{Wasserstein Space}\label{sec:gradient_wasserstein}
For distributional responses (Experiments~\ref{case:distn_lowdim2} and~\ref{case:distn_highdim}), the response space is the $2$-Wasserstein space $\mathcal{W}_2$ of univariate probability measures. 
Let $Q_i = F_i^{-1}$ denote the quantile function of the response $Y_i$, where $F_i$ is its cumulative distribution function. 
Each response is represented by $Q_i$ evaluated on a common grid $0 \le t_1 < \cdots < t_N \le 1$, giving $\vq_i = (Q_i(t_1),\ldots,Q_i(t_N))\tps\in\real^N$; the grid used in each experiment is specified in Section~\ref{sec:empircal}. 
Denote unconstrained weighted average by $\bar{\vq}_j = \sum_{i=1}^n \widehat{w}_{ij}\vq_i$ with $\widehat{w}_{ij}$ defined in Eq.~\eqref{eq:gfrwtij}. 
Since the weights $\widehat{w}_{ij}$ can be negative while summing to one, 
entries in $\bar{\vq}_j$ need not be non-decreasing, and hence need not correspond to a valid quantile function.
The quantile function corresponding to the weighted Fr\'echet mean $\predFNNhat(\vX_j;\wtset_{\nlyr})$ in $\mathcal{W}_2$ is instead obtained by projecting onto the isotonic cone $\mathcal{Q} = \{\vq\in\real^N: q_1\le\cdots\le q_N\}$ \citep{pete:19},
\bal\label{eq:pava_proj}
\hat{\vq}_j = \Pi_{\mathcal{Q}}(\bar{\vq}_j) = \argmin_{\vq\in\mathcal{Q}} \|\vq-\bar{\vq}_j\|^2 ,
\eal
which is well defined since $\mathcal{Q}$ is closed and convex.

The training loss is evaluated at the exact projected prediction $\hat{\vq}_j = \Pi_{\mathcal{Q}}(\bar{\vq}_j)$, computed by the pool adjacent violators algorithm \citep[PAVA;][]{barl:72}.
From the KKT conditions of Eq.~\eqref{eq:pava_proj}, the solution partitions $\{1,\ldots,N\}$ into level sets $B_1, \dots, B_K$ such that for $\bar{q}_{j,\ell}\in B_{k},\bar{q}_{j,\ell'}\in B_{k'}$, one has $\bar{q}_{j,\ell} <\bar{q}_{j,\ell'}$ if $k<k'$. 
On each level set $B_k$, define
\[
\hat{q}_{j,\ell} = |B_k|^{-1}\sum_{\ell'\in B_k}\bar{q}_{j,\ell'} \quad \text{for } \ell\in B_k.
\]
The map $\Pi_{\mathcal{Q}}$ is piecewise linear, and wherever the block structure is locally constant it is differentiable with block-diagonal Jacobian
\bal\label{eq:pava_jacobian}
\frac{\partial\hat{q}_{j,\ell}}{\partial\bar{q}_{j,\ell'}} =
|B_k|^{-1}\,\mathbf{1}\{\ell,\ell'\in B_k\text{ for some }k\} .
\eal
The block structure changes only on a set of Lebesgue measure zero, on which Eq.~\eqref{eq:pava_jacobian} is an element of the Clarke generalized Jacobian; we use it as the backward pass throughout.
We implement this as a custom \texttt{torch.autograd.Function} whose forward pass runs PAVA and whose backward pass applies Eq.~\eqref{eq:pava_jacobian}.

\subsection{Graph Laplacian Space}\label{sec:gradient_laplacian}

For network-valued responses (Experiment~\ref{case:network}), the response space 
$\mathcal{L}_\nnode$ consists of graph Laplacian matrices of undirected networks with 
$\nnode$ nodes, endowed with the Frobenius metric. 
A symmetric matrix $\mL = (L_{k\ell})\in\real^{\nnode\times\nnode}$ belongs to $\mathcal{L}_\nnode$ if and only if 
\bal\label{eq:laplacian_constraints}
L_{k\ell}\le 0\ \text{ for all }k\ne\ell, \quad \text{and}\quad \mL\bm1_\nnode = \bm0_\nnode ,
\eal
so that $\mathcal{L}_\nnode$ is a polyhedral convex cone. 
Let $\mL_i$ denote the graph Laplacian of $Y_i$ for $i=1,\dots,n$. 
Fix a $j\in\{1,\dots,n\}$ arbitrarily. 
The unconstrained weighted average $\bar{\mL}_j=\sum_{i=1}^n\widehat w_{ij}\mL_i$ is symmetric 
but need not satisfy the sign constraints in Eq.~\eqref{eq:laplacian_constraints} when some weights are negative, 
and the weighted Fr\'echet mean $\predFNNhat(\vX_j;\wtset_{\nlyr})$ over $\mathcal{L}_\nnode$ is the projection 
\bal\label{eq:laplacian_proj}
\hat\mL_j = \Pi_{\mathcal{L}_\nnode}(\bar \mL_j)
 = \argmin_{\mL\in\mathcal{L}_\nnode}\tfrac12\|\mL-\bar\mL_j\|_{\frob}^2 , 
\eal
which is well defined since $\mathcal{L}_\nnode$ is closed and convex. 

Unlike the isotonic projection in Section~\ref{sec:gradient_wasserstein}, which is obtained by averaging within the level sets identified by PAVA, 
Eq.~\eqref{eq:laplacian_proj} has no comparably explicit solution and is instead characterized by its optimality conditions. 
As a quadratic program problem, it has $\nnode(\nnode+1)/2$ free primal coordinates, together with $\nnode$ multipliers for the row-sum constraints and $\nnode(\nnode-1)/2$ multipliers for the sign constraints, and in this form it can be passed to a general-purpose solver. 
Because the sign constraints act on the off-diagonal entries separately, however, both the primal variable and its sign multipliers can be eliminated in closed form once the row-sum multipliers $\bm\lambda\in\real^\nnode$ are fixed, leaving a system of size $\nnode$. 
We solve the problem in this reduced form. The resulting projection map is piecewise affine and hence differentiable almost everywhere; 
whenever the active set is locally unchanged, its ordinary derivative is available at the cost of a single $\nnode\times\nnode$ factorization. 

\subsubsection{Reduction to the Row-Sum Multipliers}
Let $\operatorname{diag}(\mbf M)$ be the vector of diagonal entries of a
matrix $\mbf M$ 
and $\odot$ denote the entrywise product, 
and set
\bal\label{eq:lap_residual}
\mE(\bm\lambda) = \bar \mL_j - \tfrac12\bigl(\bm\lambda\bm1_\nnode\tps 
   + \bm1_\nnode\bm\lambda\tps \bigr)
\quad \text{with }
E_{k\ell}(\bm\lambda) = \bar L_{j,k\ell} - \tfrac12(\lambda_k+\lambda_\ell).
\eal
Attaching multipliers $\bm\lambda$ to the row-sum constraints $\mL\bm1_\nnode = \bm0_\nnode$ and $\mu_{k\ell}\ge0$ to the sign constraints $L_{k\ell}\le 0$ for all $k\ne \ell$, 
the Lagrangian is
\bal\label{eq:lap_lagrangian}
\mathscr{L}(\mL;\bm\lambda,\{\mu_{k\ell}\}_{k\ne\ell})
  = \tfrac{1}{2}\|\mL-\bar\mL_j\|_F^2
    + \bm\lambda\tps\mL\bm1_\nnode
    + \sum_{k\ne\ell}\mu_{k\ell}L_{k\ell}.
\eal
Stationarity of the Lagrangian $\partial\mathscr{L}/\partial\mL=\bm{0}_{q\times q}$ gives $\hat L_{j,kk}=\bar L_{j,kk}-\lambda_k$ for $k=1,\dots,q$ and $\hat L_{j,k\ell}=E_{k\ell}(\bm\lambda)-\mu_{k\ell}$ for any distinct $k,\ell\in\{1,\dots,q\}$, 
the latter after dividing by two to account for each off-diagonal entry appearing in both the $(k,\ell)$ and the $(\ell,k)$ slot of the symmetric matrix $\mL$. 
For a fixed off-diagonal pair, complementary slackness $\mu_{k\ell}\hat L_{j,k\ell}=0$ forces one of the two unknowns to vanish. 
Setting $\mu_{k\ell}=0$ gives $\hat L_{j,k\ell}=E_{k\ell}(\bm\lambda)$, which is primal feasible exactly when $E_{k\ell}(\bm\lambda)\le0$; 
setting $\hat L_{j,k\ell}=0$ gives $\mu_{k\ell}=E_{k\ell}(\bm\lambda)$, which is dual feasible exactly when $E_{k\ell}(\bm\lambda)\ge0$. 
The two ranges cover the real line and the two branches coincide where they overlap, so each off-diagonal entry is determined for every $\bm\lambda$ as the projection of $E_{k\ell}(\bm\lambda)$ onto $(-\infty,0]$:
\bal\label{eq:lap_primal_from_dual}
\hat L_{j,k\ell} = \min\{E_{k\ell}(\bm\lambda),\,0\}\quad(k\ne\ell),
\qquad
\hat L_{j,kk} = \bar L_{j,kk}-\lambda_k .
\eal
Since all constraints in Eq.~\eqref{eq:laplacian_constraints} are affine and the objective is strictly convex, these conditions are not merely necessary but also sufficient. 
Substituting Eq.~\eqref{eq:lap_primal_from_dual} into the row-sum constraint leaves the $\nnode$-dimensional equation
\bal\label{eq:lap_dual_eqn}
F(\bm\lambda) = \operatorname{diag}(\bar \mL_j)-\bm\lambda
+\bigl(A_{\mathcal{E}(\bm\lambda)}\odot R(\bm\lambda)\bigr)\bm1_\nnode \;=\;\bm0_\nnode,
\eal
where $\mathcal{E}(\bm\lambda)=\{(k,\ell): k\ne\ell,\ E_{k\ell}(\bm\lambda)<0\}$ is the support set 
and $A_{\mathcal{E}(\bm\lambda)}$ is its symmetric binary adjacency matrix. 
Solving Eq.~\eqref{eq:lap_dual_eqn} and plugging into Eq.~\eqref{eq:lap_primal_from_dual} returns the projection. 
Since the objective in Eq.~\eqref{eq:laplacian_proj} is strictly convex and $\mathcal{L}_q$ is closed and convex, 
$\hat\mL_j$ is unique by the Hilbert-space projection theorem. 
The dual solution $\bm\lambda$ is then also unique: stationarity of the Lagrangian on the diagonal gives 
$\lambda_k = \bar L_{j,kk} - \hat L_{j,kk}$, which is determined once $\hat\mL_j$ is fixed. 

\subsubsection{Semismooth Newton Solver}
Eq.~\eqref{eq:lap_dual_eqn} is piecewise affine in $\bm\lambda$, the pieces being the regions on which $\mathcal{E}(\bm\lambda)$ is constant, and hence semismooth. 
On the interior of any such region, differentiating through Eq.~\eqref{eq:lap_residual} gives
$\partial F/\partial\bm\lambda=-H(\mathcal{E})$ with 
\bal\label{eq:lap_newton_matrix}
H(\mathcal{E})=\mI_\nnode+\tfrac12\bigl(\mD_{\mathcal{E}}+\mA_{\mathcal{E}}\bigr) 
\quad \text{and}\quad
[\mD_{\mathcal{E}}]_{kk}=\bigl[\mA_{\mathcal{E}}\bm1_\nnode\bigr]_k.
\eal
The matrix $\mD_{\mathcal{E}}+\mA_{\mathcal{E}}$ is the signless Laplacian of the support graph and is positive semidefinite, 
because $\bm x\tps (\mD_{\mathcal{E}}+\mA_{\mathcal{E}})\bm x
=\tfrac12\sum_{k\ne\ell}[\mA_{\mathcal{E}}]_{k\ell}(x_k+x_\ell)^2$, 
so that $H(\mathcal{E})\succeq I_\nnode$ and every element of the generalized Jacobian of $F$ is invertible. 
Starting from $\bm\lambda=\bm0_\nnode$, we iterate the semismooth Newton step 
$\bm\lambda\leftarrow\bm\lambda+H(\mathcal{E}(\bm\lambda))^{-1}F(\bm\lambda)$, 
computed by a Cholesky factorization of an $\nnode\times\nnode$ matrix, until $\|F(\bm\lambda)\|_\infty$ falls below a tolerance. 
Because $F$ is piecewise affine, the iteration terminates one step after the support set stabilizes, and in our experiments a handful of iterations suffices. 

\subsubsection{IFT-Based Differentiable Training Loss} 
The empirical \F risk in Eq.~\eqref{eq:dfnn_risk_emp} is evaluated at the exact projection $\hat \mL_j$, so the backward pass must differentiate the map $\bar \mL_j\mapsto\hat \mL_j$ defined by Eqs.~\eqref{eq:lap_primal_from_dual}--\eqref{eq:lap_dual_eqn}. 
Although both $F$ and the resulting projection map are only piecewise affine globally, they are affine on each region where the support set is fixed. Accordingly, we first derive the ordinary derivative away from active-set boundaries. 
Suppose strict complementarity holds at the solution $\bm\lambda^\ast$, that is, $E_{k\ell}(\bm\lambda^\ast)\ne0$ for all $k\ne\ell$. 
Then the support set is locally unchanged, and the implicit function theorem (IFT) \citep[Theorem~9.28]{rudi:76} applies. 
Each $E_{k\ell}$ is continuous, so the sign pattern and hence the support set $\mathcal{E}^\ast=\mathcal{E}(\bm\lambda^\ast)$ is constant on a neighbourhood of $(\bm\lambda^\ast,\bar L_j)$, 
on which $F$ is affine in both arguments with invertible partial Jacobian $-H(\mathcal{E}^\ast)$. 
The solution map is therefore differentiable at $\bar \mL_j$, and differentiating $F=\bm0_\nnode$, using $\diffop E_{k\ell}=\diffop \bar L_{j,k\ell}$ at fixed $\bm\lambda$, gives
\bal\label{eq:lap_ift}
H(\mathcal{E}^\ast)\,\diffop \bm\lambda
 = \operatorname{diag}(\diffop \bar \mL_j)+\bigl(A_{\mathcal{E}^\ast}\odot \diffop \bar \mL_j\bigr)\bm1_\nnode .
\eal
Differentiating Eq.~\eqref{eq:lap_primal_from_dual} on the same neighbourhood gives 
$\diffop \hat L_{j,kk}=\diffop \bar L_{j,kk}-\diffop \lambda_k$, together with 
$\diffop \hat L_{j,k\ell}=\diffop \bar L_{j,k\ell}-\tfrac12(\diffop \lambda_k+\diffop \lambda_\ell)$ for 
$(k,\ell)\in\mathcal{E}^\ast$ and $\diffop \hat L_{j,k\ell}=0$ for the remaining off-diagonal entries.

Combining the two relations yields the vector--Jacobian product in closed form. 
Let $\mG=\partial\efrisk/\partial\hat \mL_j$ be the incoming cotangent and $\mG^{\mathrm{sym}}=\tfrac12(\mG+\mG\tps )$, 
where $\efrisk = \efrisk\{\predFNNhat(\cdot;\wtset_{\nlyr})\}$ is the empirical \F risk in Eq.~\eqref{eq:dfnn_risk_emp}. 
Contracting $\mG$ with the primal differential in the direction of $\bm\lambda$ and solving the adjoint of Eq.~\eqref{eq:lap_ift} gives 
\bal\label{eq:lap_adjoint}
\bm h=-\diag(\mG)-\bigl(\mA_{\mathcal{E}^\ast}\odot \mG^{\mathrm{sym}}\bigr)\bm1_\nnode,
\qquad
\bm v=H(\mathcal{E}^\ast)^{-1}\bm h ,
\eal
and propagating back to $\bar \mL_j$, 
\bal\label{eq:lap_vjp}
\frac{\partial\efrisk}{\partial\bar \mL_j}
= \bigl(\mA_{\mathcal{E}^\ast} + \mI_\nnode\bigr)\odot\bigl(\mG+\bm v\bm1_\nnode\tps \bigr),
\eal
symmetrized to respect the symmetry of $\bar \mL_j$. 
Each entry of Eq.~\eqref{eq:lap_vjp} collects the direct dependence of $\hat \mL_j$ on $\bar \mL_j$ and the indirect dependence through $\bm\lambda$; inactive off-diagonal entries receive no contribution because $\hat L_{j,k\ell}\equiv0$ there. 
The backward pass reuses the Cholesky factor of $H(\mathcal{E}^\ast)$ from the forward solve and costs one additional $\nnode$-dimensional triangular solve. 

Finally, the set of $\bar \mL_j$ at which strict complementarity fails is negligible. 
For each candidate support set $\mathcal{E}$, the region of $\bar \mL_j$ producing it is a polyhedron, 
since there $\bm\lambda^\ast$ is the affine function of $\bar \mL_j$ obtained 
by solving the linear system with fixed matrix $H(\mathcal{E})$, and there are finitely many such regions. 
Within each region $\bar \mL_j\mapsto E_{k\ell}(\bm\lambda^\ast(\bar \mL_j))$ is affine and nonconstant, 
its derivative in the symmetric direction supported on $(k,\ell)$ being $1$ when $(k,\ell)\notin\mathcal{E}$ and $1-\tfrac12(\bm e_k+\bm e_\ell)\tps H(\mathcal{E})^{-1}(\bm e_k+\bm e_\ell)>0$ otherwise, 
where positivity follows from $H(\mathcal{E})\succeq \mI_\nnode$ together with $[\mA_{\mathcal{E}}]_{k\ell}=1$. 
Its zero set is therefore a hyperplane, and the degenerate set, being contained in a finite union of hyperplanes, has Lebesgue measure zero. 
There Eq.~\eqref{eq:lap_vjp} remains an element of the Clarke generalized Jacobian, and we use it as the backward pass throughout, exactly as in
Section~\ref{sec:gradient_wasserstein}.

\subsection{Aitchison Space}\label{sec:gradient_compositional}
For compositional responses (Experiment~\ref{case:employment_composition}), the
response space is the $(q-1)$-simplex
$\Delta^{q-1} = \{\va\in[0,1]^q:\sum_{k=1}^q a_k=1\}$ endowed with the
Aitchison distance $d_A(\va,\vb) = \|\mathrm{clr}(\va)-\mathrm{clr}(\vb)\|$, for any $\va,\vb\in\Delta^{q-1}$ \citep{aitc:86}. 
Here, 
\bal\label{eq:clr_defn}
\mathrm{clr}(\va) = \log(\va) - \frac{1}{q}\bm{1}_q\tps \log(\va)\cdot\bm{1}_q,
\eal
is the centered log-ratio (CLR) transformation, an isometry from $(\Delta^{q-1},d_A)$ onto
the hyperplane $(H_q,\|\cdot\|)$, 
where $\log(\va)$ denotes the entry-wise logarithmic transformation 
and $H_q = \{\vv\in\real^q: \sum_{k=1}^q v_k=0\}$. 
Since $H_q$ is a linear subspace of $\real^q$, it is closed under linear combinations; 
specifically, $\bar{\vz}_j = \sum_{i=1}^n \widehat{w}_{ij}\,\mathrm{clr}(Y_i)$ lies in $H_q$, for any $j=1,\dots,n$. 
The weighted Fr\'echet mean $\predFNNhat(\vX_j;\wtset_{\nlyr})$ in Aitchison simplex $\Delta^{q-1}$ is therefore available in closed form:
\bal\label{eq:clr_frechet}
\hat{\mathbf{a}}_j = \mathrm{clr}^{-1}(\bar{\vz}_j)
 = \exp(\bar{\vz}_j)\big/\textstyle\sum_{k=1}^q \exp(\bar z_{j,k}) ,
\eal
with no projection step required.
The derivative
$\partial\bar{\vz}_j/\partial\widehat{w}_{ij} = \mathrm{clr}(Y_i)$
is immediate, and the remaining maps in Eq.~\eqref{eq:clr_frechet} are smooth, so the
gradient is obtained by standard automatic differentiation.

\subsection{Sphere}\label{sec:gradient_sphere}
The responses in Experiment~\ref{case:energy_sphere} are essentially spherical: each observation is a vector of $q+1$ non-negative parts summing to one, 
mapped to the unit sphere $\sphere^{q}=\{\vv\in\real^{q+1}:\|\vv\|=1\}$ by the component-wise square root transformation \citep{scea:14}, which lie in the closed positive orthant $\sphere^{q}_+=\{\vv\in\sphere^{q}: v_1,\ldots,v_{q+1}\ge0\}$. 
The orthant is geodesically convex and contained in the geodesic ball centered at $(q+1)^{-1/2}\bm 1_{q+1}$ of radius $\arccos\{(q+1)^{-1/2}\}<\pi/2$. 
Let $Y_i$ denote the square root transformed composition, 
and the response space is $\sphere^{q}$ equipped with the geodesic distance $d_g(\vv,\vv')=\arccos(\vv\tps \vv')$. 

Unlike the cases above, $\sphere^{q}$ is not isometric to a subset of a Hilbert
space, and the objective function $\ofFNNhat(\cdot,\vx;\wtset_{\nlyr})$ in Eq.~\eqref{eq:dfnn_output_layer_emp}  for $\vx=\vX_j$ becomes 
\bal\label{eq:wfm_sphere}
\ofFNNhat_j(\vu) = \sum_{i=1}^n \widehat{w}_{ij}\,d_g^2(Y_i,\vu)
\quad \text{for } \vu\in\sphere^{q}
\eal
and has no minimizer available in closed form.  
Denote 
\[
\hat{\vu}_j = \predFNNhat(\vX_j;\wtset_{\nlyr}) 
= \argmin_{\vu\in\sphere^{q}}\ofFNNhat_j(\vu).
\]
We compute $\hat{\vu}_j$ by the
Riemannian gradient descent (RGD) \citep{karc:77} detailed belows, 
of which the solution satisfies $\nabla\ofFNNhat_j(\hat{\vu}_j)=\bm 0$.  
When the weights are non-negative the minimizer is unique; 
otherwise, $\ofFNNhat_j$ need not be geodesically convex and RGD returns a local minimizer.

\subsubsection{Karcher Flow}
For $\vu\in\sphere^{q}$, the tangent space at $\vu$ is $T_{\vu}\sphere^{q}=\{\vv\in\real^{d+1}:\vv\tps \vu=0\}$, a $q$-dimensional linear subspace of $\real^{q+1}$. 
The exponential map
$\exp_{\vu}\colon T_{\vu}\sphere^{q}\to\sphere^{q}$ and the logarithmic map $\log_{\vu}\colon \sphere^{q}\setminus\{-\vu\}\to T_{\vu}\sphere^{q}$ are defined as 
\bal\label{eq:sphere_log}
\exp_{\vu}(\vz) = \cos\|\vz\|\,\vu + \sin\|\vz\|\,\frac{\vz}{\|\vz\|} 
\quad\text{and}\quad 
\log_{\vu}(\vv)
= \frac{\arccos(\vv\tps \vu)}
{\|\vv-(\vv\tps \vu)\vu\|}
\bigl(\vv-(\vv\tps \vu)\vu\bigr),
\eal
with $\log_{\vu}(\vu)=\bm 0$, $\exp_{\vu}(\bm 0)=\vu$, and $\|\log_{\vu}(\vv)\|=d_g(\vu,\vv)$. 
As $\nabla_{\vu}\,d_g^2(\vv,\vu)=-2\log_{\vu}(\vv)$, Riemannian gradient descent on $\tfrac12\ofFNNhat_j(\cdot)$ with step size $\alpha>0$ moves along the geodesic in the direction 
$\vz^{(k)}=\alpha\sum_{i=1}^n\widehat{w}_{ij}\log_{\vu^{(k)}}(Y_i)
\in T_{\vu^{(k)}}\sphere^{q}$, giving the iterative updates 
\bal\label{eq:karcher_update}
\vu^{(k+1)} = \exp_{\vu^{(k)}}(\vz^{(k)}),
\quad \text{for } k=0,1,\ldots,
\eal
which starts at the normalized Euclidean average 
$\vu^{(0)}=\sum_{i=1}^{n}\widehat{w}_{ij}Y_i/\|\sum_{i=1}^{n}\widehat{w}_{ij}Y_i\|$ 
and stops once $\|\vz^{(k)}\|/\alpha$ falls below a pre-specified tolerance. 

\subsubsection{Hessian and Conditions}
Write $\rho_{ij}=d_g(Y_i,\hat{\vu}_j)$, 
let $\ve_{ij}=\log_{\hat{\vu}_j}(Y_i)/\rho_{ij}$ for $\rho_{ij}\neq0$ be the unit tangent vector towards $Y_i$, 
and let $\mP=\mI_{q+1}-\hat{\vu}_j\hat{\vu}_j\tps $ be the orthogonal projector onto $T_{\hat{\vu}_j}\sphere^{q}$. 
The covariant Hessian of $\tfrac12\ofFNNhat_j(\cdot)$ at $\hat{\vu}_j$ is given by 
\bal\label{eq:sphere_hessian}
\mH_j = \Bigl(\sum_{i=1}^n \widehat{w}_{ij}\,\rho_{ij}\cot\rho_{ij}\Bigr)\mP
+ \sum_{i:\rho_{ij}\neq0} \widehat{w}_{ij}\bigl(1-\rho_{ij}\cot\rho_{ij}\bigr)
   \ve_{ij}\ve_{ij}\tps ,
\eal
a symmetric linear map on $T_{\hat{\vu}_j}\sphere^{q}$, with $\rho\cot\rho$ read as its limit $1$ at $\rho=0$. 
We assume with probability $1$: 
\begin{enumerate}[label=(S\arabic*),leftmargin=*]
\item\label{cond:sphere_cut} 
There exists $\varepsilon>0$ such that $\max_{1\le i\le n}\rho_{ij}\le\pi-\varepsilon$ for every $j=1,\ldots,n$.
\item\label{cond:sphere_hess} 
For every $j=1,\ldots,n$, $\mH_j$ is nonsingular on $T_{\hat{\vu}_j}\sphere^{q}$. 
\end{enumerate}
Since $\log_{\vu}(\vv)$ is smooth in $\vu$ away from $-\vv$,
\ref{cond:sphere_cut} places $\hat{\vu}_j$ in a neighborhood on which
$\nabla\ofFNNhat_j$ is continuously differentiable, and applying the implicit function theorem \citep[Theorem~9.28]{rudi:76} 
to $\nabla\ofFNNhat_j(\hat{\vu}_j)=\bm 0$ yields 
\bal\label{eq:sphere_ift}
\frac{\partial\hat{\vu}_j}{\partial\widehat{w}_{ij}}
 = \mH_j^{-1}\log_{\hat{\vu}_j}(Y_i)
 \in T_{\hat{\vu}_j}\sphere^{q},
\eal
the inverse taken on $T_{\hat{\vu}_j}\sphere^{q}$, which is evaluated in the backward pass of a custom automatic differentiation rule. 
Both conditions were monitored in Experiment~\ref{case:energy_sphere}: 
$\max_{i}\rho_{ij}$ never exceeded $2.15$ rad ($123^\circ$) throughout, 
so \ref{cond:sphere_cut} held with $\varepsilon\approx0.996$; 
and $\mH_j$, being symmetric, had smallest tangential eigenvalue at least $0.504$ over all $j$ during $100$ Monte Carlo runs, so that $\mH_j\succ0$ and each $\hat{\vu}_j$ was a strict local minimizer of $\ofFNNhat_j(\cdot)$. 

\end{document}